\documentclass[12pt,a4paper]{article}

\usepackage{amsmath,amssymb,graphicx}
\usepackage[english]{babel}
\usepackage{bm,url,float}
\usepackage[a4paper,text={17cm,25.5cm},centering]{geometry}
\usepackage[compact,small]{titlesec}
\usepackage{lmodern}
\usepackage{tabularx}
\usepackage[scientific-notation=true]{siunitx}
\usepackage{natbib}
\usepackage{nicefrac}
\usepackage{ltablex}
\usepackage{booktabs,fancyvrb}
\usepackage{array}
\usepackage{multirow,footnote}
\usepackage{color,rotating}
\usepackage{lscape}
\usepackage{tikz}
\usepackage[hidelinks]{hyperref}
\usepackage{authblk}
\usepackage{scrextend}
\usepackage{xcolor}
\usepackage{xurl}
\usepackage{longtable,caption}
\usepackage{adjustbox}
\usepackage[doublespacing]{setspace}
\usepackage{subcaption}
\usepackage{scalerel,stackengine}

\usepackage{enumitem}

\newcolumntype{L}[1]{>{\raggedright\arraybackslash}m{#1}}
\newcolumntype{C}[1]{>{\centering\arraybackslash}m{#1}}
\newcolumntype{R}[1]{>{\raggedleft\arraybackslash}m{#1}}

\stackMath
\newcommand\reallywidehat[1]{%
\savestack{\tmpbox}{\stretchto{%
\scaleto{%
\scalerel*[\widthof{\ensuremath{#1}}]{\kern-.6pt\bigwedge\kern-.6pt}%
{\rule[-\textheight/2]{1ex}{\textheight}}%
}{\textheight}%
}{0.5ex}}%
\stackon[1pt]{#1}{\tmpbox}%
}

\graphicspath{{Figures/}}

\begin{document}

\begin{titlepage}
    \title{From Tone to Trajectory: Continuous Sentiment and the Shape of Monetary Policy Communication}
\author[1,2,3]{MARTIN FELDKIRCHER}
\author[4]{MÁRTON KARDOS}
\author[4]{KRISTOFFER LAIGAARD NIELBO\thanks{Martin Feldkircher \texttt{martin.feldkircher@da-vienna.ac.at} (corresponding author), Márton Kardos, \texttt{martonkardos@cas.au.dk}, and Kristoffer Laigaard Nielbo \texttt{kln@cas.au.dk}. Any views expressed in this paper represent those of the authors only and do not necessarily coincide with those of the Oesterreichische Nationalbank or the Eurosystem. We would like to thank Kilian Rieder and participants of the ZEW workshop in Mannheim, and an internal seminar at the Oesterreichische Nationalbank for helpful comments.}}

\affil[1]{Vienna School of International Studies (DA)}
\affil[2]{Oesterreichische Nationalbank (OeNB)}
\affil[3]{Centre for Applied Macroeconomic Analysis (CAMA)}
\affil[4]{School of Culture and Society - Center for Humanities Computing (AU)}

    \date{\today}
    \maketitle

    \begin{abstract}
    \begin{singlespace}
    \noindent
    Central bank press conferences are not merely information releases --- they are structured narratives. We study whether the \emph{shape} of sentiment within a statement, not just its average tone, carries policy-relevant signals. Constructing  sentiment arcs for ECB and Fed press conferences along three dimensions --- monetary stance, economic outlook, and uncertainty --- we assess their predictive content for policy rate changes, inflation expectations, and forecaster disagreement. Our findings show that arc shape robustly predicts rate decisions beyond lexicon-based benchmarks at both institutions --- it is not merely whether a statement sounds hawkish or economically optimistic on average, but how these sentiments are sequenced and emphasized across the statement, that carries the policy signal. Arc features also shape how professional forecasters update inflation expectations and how much they disagree, pointing to a receiver-side effect distinct from the direct policy signal. These findings suggest that communication design --- the sequencing and emphasis of policy language across a statement --- is a first-order feature of the policy signal, not a second-order refinement.
    \end{singlespace}
    \end{abstract}

    \bigskip
    \begin{flushleft}
    \end{flushleft}

    \setcounter{page}{0}
    \thispagestyle{empty}
\end{titlepage}
\pagebreak \newpage

\section{Introduction}\label{sec:intro}

Over the last two decades, communication has become an increasingly important aspect of monetary policy \citep{Blinder2008, Blinder2024}. Central banks have moved away from deliberate opacity toward active management of expectations, investing heavily in forward guidance, structured press conferences, and clearly worded policy statements \citep{Blinder2008}. The European Central Bank (ECB) and the US Federal Reserve (Fed) are two prominent examples. Following each Governing Council meeting, the ECB President delivers a prepared introductory statement before opening the floor to journalists' questions; the statement and the press conference together constitute distinct information events, each with its own effect on financial markets \citep{Altavilla2019, EhrmannFratzscher2009}. The Fed similarly publishes a carefully worded statement after every FOMC meeting. These documents are crafted with awareness that every word will be scrutinised by market participants \citep{Hubert2021}.

A sizeable empirical literature documents that this communication content matters: ECB press conferences provide substantial additional information to financial markets beyond what is contained in the rate decision itself \citep{RosaVerga2007, EhrmannFratzscher2009, ConradLamla2010, Parle2022,Gardner2022}, the tone of CB statements shapes inflation expectations \citep{ChoJung2026, ChoRho2026, Montes2016, Hoffmann2026}, and communication content helps predict future policy decisions \citep{Kanelis2024, Baranowski2021, Hubert2021}. Much of this evidence reduces a policy statement to a single aggregate tone or sentiment measure --- a scalar capturing the overall hawkish or dovish inclination of the text \citep{Picault:2017, Baranowski2021}. While this approach has proven productive, it discards the internal structure of the communication: how sentiment evolves \emph{within} a statement, where it rises or falls, whether it accelerates or reverses, and how the statement ends.

We argue that this internal structure --- the \emph{sentiment arc} --- carries information not captured by the aggregate tone. A sentiment arc represents the sequential evolution of sentiment scores across a document. Deriving sentiment arcs instead of a single aggregate sentiment score for a document can provide more granular information about local variation in affective tone. Our motivation draws on two strands of work. First, research in computational narratology shows that emotional arcs of stories are not arbitrary but dominated by a small number of basic shapes, and that particular arc shapes are associated with greater audience success \citep{Reagan2016}; this regularity extends across different genres and communication contexts, suggesting that arc shape is a robust feature of structured communication \citep{Neugarten2025}.\footnote{Sentiment arc features have also been shown to carry information about the perceived quality of literary narratives beyond what average sentiment conveys \citep{Bizzoni2022}.} Second, evidence from psychology documents a peak-end bias: retrospective evaluations of sequential experiences are systematically governed by the peak and the endpoint, not the time-averaged experience \citep{Mueller2019}. If market participants process policy documents in a similar fashion, the shape of the arc --- and especially how the statement ends --- may convey disproportionate policy-relevant signals beyond what aggregate tone captures.

To investigate this, we construct word-level sentiment scores for the full corpus of ECB introductory statements and FOMC statements along three conceptually distinct dimensions: (i) \emph{monetary sentiment} (hawkish vs.\ dovish), capturing the direction of the signalled policy intention, (ii) \emph{economic sentiment} (positive vs.\ negative), capturing the central bank's assessment of underlying economic conditions --- both motivated and assessed in \citet{Picault:2017} --- and (iii) \emph{uncertainty sentiment}, measuring the degree of epistemic uncertainty conveyed by the statement. Scores are then derived using Concept Vector Projection (CVP), a method proposed by \citet{Lyngbaek2025} that projects sentence embeddings from a large language model onto a direction in embedding space defined by manually curated seed phrases. Relative to bag-of-words dictionaries \citep{Picault:2017}, fine-tuned classifiers \citep{Shah2023}, and large language model-based scoring \citep{HansenKazinnik2024, SilvaMoriyaVeyrune2025}, CVP provides continuous scores in a common, context-sensitive embedding space without requiring model fine-tuning or task-specific prompting.

Our main results are as follows. Arc shape robustly predicts policy rate changes at both the ECB and the Federal Reserve, beyond what lexicon-based benchmark measures capture. A common core of findings holds across institutions: statements that are hawkish or become progressively more hawkish over their course, and statements in which the economic outlook is positive or concentrated positively in the analytical middle, signal subsequent rate increases. Beyond this common core, institution-specific features emerge. At the ECB, the persistence of the monetary sentiment arc --- how self-similar the hawkish--dovish signal is across the statement --- carries additional predictive content, with less persistent arcs signalling rate decreases. At the Fed, the uncertainty dimension matters: statements whose expressed uncertainty rises toward the close are indicative of rate cuts.  On the receiver side, arc features predict how professional forecasters update inflation expectations and how much they disagree, with the relevant arc dimension differing by institution: monetary arc features matter at the ECB, economic arc features at the Fed. Finally, crisis episodes generate heterogeneous arc--rate relationships: the ECB's arc structure proves stable across regimes, while the standard relationship between a hawkish arc and rate increases reverses sharply at the Fed during the GFC and COVID periods.

These results contribute to the literature on central bank communication \citep{Blinder2008, EhrmannFratzscher2009, Ehrmann2024, Wabitsch2025} in two ways. Empirically, we show that the arc structure of policy statements encodes information about future rate decisions that aggregate sentiment scores miss. Conceptually, we establish a link between the computational narratology literature on emotional arcs \citep{Reagan2016} and the central bank communication literature, framing the rhetorical architecture of policy documents as a deliberate signal of communication design.

The remainder of the paper is organised as follows. Section~\ref{sec:data} describes the corpus of ECB and FOMC statements and introduces the methodology. Section~\ref{sec:descriptive} presents the full-sample sentiment arcs and validates the scoring approach. Section~\ref{sec:results} reports results from predicting rate changes and inflation expectations. Finally, Section~\ref{sec:conclusions} concludes.

\section{Data and Methodology}\label{sec:data}

For this study, we utilize data from the press conferences of the European Central Bank (ECB). The ECB's Governing Council convenes regularly to decide on monetary policy, with outcomes communicated through two steps \citep{Altavilla2019}.
First, at 13:45 Central European Time (CET), the ECB releases a concise statement regarding its monetary policy decision (MPD). This statement is brief and contains limited textual content.
Second, at 14:30 CET, the ECB President delivers an \emph{introductory statement} during a press conference. This carefully crafted document informs the public about the rationale behind the interest rate choices, presents the ECB's perspective on the economic situation, and provides insights into its future conduct. From 1999 to 2003, each statement followed a two-pillar template: it opens with the policy decision, proceeds through the monetary analysis (covering monetary dynamics and credit conditions) to the 
economic analysis (covering the inflation outlook and near-to-medium-term risks to price stability). The ECB's 2003 strategy review reversed this ordering: from 2003 to July 2013 the decision is followed by economic analysis, which is followed by the monetary analysis and finally a cross-check \citep{HolmHadulla2021}. Since 2013, to this structure an explicit forward guidance chapter is added. Following another strategy review, from July 2021 the statement contains the following sections, rate decision and inflation projections; economic activity, inflation, risk assessment, monetary conditions and conclusions. This fixed sequential structure, combined with the need to present a comprehensive analytical narrative at each press conference, produces documents that are substantially longer than FOMC statements. The introductory statement lasts approximately fifteen minutes and is followed by a forty-five-minute session of questions and answers.  We obtain the press conference texts directly from the ECB's website (\url{https://www.ecb.europa.eu/press/pressconf/html/index.en.html}) using automated web scraping.
Our raw text data for the ECB contains 269 statements from the period 1998-06-09 to 2025-06-06. For the predictive regressions, carried out in Section \ref{sec:results} we also obtain policy rates. These are collected from the BIS data base. For the euro area and from 12-09-2024 on, the policy rate no longer constitutes the main refinancing rate (MRO) but the deposit facility rate (DPR). The matched sample of text statements and policy rates runs from 1999-01-01 to 2025-06-06 and contains 262 observations.  

For the United States, we use the statements released by the Federal Open Market Committee (FOMC) of the Federal Reserve. After each FOMC meeting, three key documents are released over time. The FOMC statement, released immediately, outlines the economic outlook, policy decision, and forward guidance. This is by far the most important document and can be compared to the introductory statements of the ECB. Following a standardized six-part template that has been stable since 2000, each statement covers: recent economic developments, the Committee's longer-run policy goals, the rate decision, forward guidance on the future rate path, factors to be assessed in upcoming decisions, and the voting record \citep{Hopper2019}. This compressed, templated format — averaging roughly 380 words — is designed to signal through word choice and emphasis within a fixed structure rather than through compositional variation.\footnote{In addition to the statement, the Fed also releases meeting minutes, published three weeks later, which summarize the main discussions and reflect an evolution in Federal Reserve communication. Finally, verbatim transcripts of the meeting are released with a five-year lag. Here, we focus on the statements and disregard information contained in the minutes due to the three-week publication lag.} The FOMC statements are obtained from \url{https://github.com/vtasca/fed-statement-scraping}.
For the US data sample, the policy rate always refers to the effective federal funds rate and the joined data set spans the period 2000-02-02 to 2025-10-29 amounting to 219 observations.

For the inflation expectations regressions in Section~\ref{sec:audience}, we additionally use professional forecasters' one-year-ahead inflation expectations from the Consensus Economics survey, conducted around the 10th of each month. The data are converted from fixed event to fixed horizon forecasts as in \citet{Siklos2013}. To control for the persistent level of inflation expectations, we include the actual year-over-year inflation rate $\Delta p^{\mathrm{yoy}}_t$: headline HICP for the euro area, obtained from Eurostat (dataset \texttt{prc\_hicp\_midx}, annual rate of change), and headline CPI for the United States, obtained from the Federal Reserve Bank of St.\ Louis FRED database (series \texttt{CPIAUCSL}).


Because the ECB statement opens with the policy decision and then presents the analytical justification through two structured pillars, while the FOMC statement builds from an economic characterization through the rate decision to forward guidance, the arc maps onto identifiable narrative phases in both cases but with different logic: for the ECB, the arc traces the progression of analytical rationale following the announced decision; for the Fed, it traces the accumulation of evidence that culminates in the decision and subsequent guidance. The difference in document length therefore reflects institutional communication design rather than informational content, and both arc structures are interpretable within their respective institutional frameworks.

\subsection{Concept Vector Projection}

In order to calculate sentiment arcs over the duration of a press conference,
we use Concept Vector Projection (CVP) \citep{Lyngbaek2025}.
In contrast to prevailing sentiment analysis pipelines, which typically treat polarity as a multi-class classification problem
  either using bag-of-words or transformer representations,
CVP relies on the linear representation hypothesis \citep{linear_representation_hypothesis} to assign continuous sentiment scores to text. Concretely, this assumes that sentences can be ordered along a single axis from negative to positive sentiment. To apply the methodology, we have to first define concepts. Following \citet{Picault:2017}, we focus on two concepts, namely monetary sentiment (hawkish positive, dovish negative) and economic sentiment (optimistic positive, pessimistic negative). As a third dimension, we add uncertainty sentiment, where uncertainty is the positive pole. Each statement can evolve in these three dimensions, word by word. Each concept is defined by a set of positive ($P^{+}$) and negative ($N^{-}$) seed phrases that have to be specified by the researcher.


Given the embedding matrices of positive seed phrases ($P^{+}$) and negative seed phrases ($N^{-}$), this difference is computed as follows:
\begin{gather}
    \overrightarrow{v} = \bar{p} - \bar{n}\text{, where } \bar{p}_j = \frac{\sum_i P^{+}_{ij}}{|P^{+}|}\text{, and } \bar{n}_j = \frac{\sum_i N^{-}_{ij}}{|N^{-}|} \\
    \hat{v} = \frac{\overrightarrow{v}}{||\overrightarrow{v}||_2}
\end{gather}
where $|P^{+}|$ and $|N^{-}|$ denote the cardinalities of the positive and negative seed sets, respectively.\\
We refer to $\hat{v}$ as the \textit{concept vector.}

To assign a sentiment score to a given document, we first encode the document using the same transformer model to obtain an embedding vector $e$, and then project it onto the concept vector using the dot product:
\begin{equation}
    s = e \cdot \hat{v}
\end{equation}

\citet{Lyngbaek2025} apply their method in a multilingual literary context and observe substantially improved alignment with human evaluation compared to baseline approaches.
CVP is particularly advantageous for our use case for several reasons:
\begin{enumerate}[noitemsep]
    \item It does not necessitate finetuning transformer models, only inference. Model training would be considerably more computationally intensive and requires access to large-scale data.
    \item CVP, thanks to neural transformer embeddings, is context sensitive  and can deal with synonymy, polysemy, and spelling errors.
    \item It allows us to compute continuous sentiment scores throughout the course of the document, thereby allowing us to compute so-called sentiment arcs.
    \item CVP's flexibility allows us to measure sentiment from multiple angles.
\end{enumerate}

To come up with potential seed phrases, we use a multi-layered process. First, we draw on the dictionary from \citet{Picault:2017} to gather candidate sentences from both the ECB and FOMC corpora. Following \citet{Lyngbaek2025}, we target 15 seed phrases per pole. In a second step, we then remove any ECB or US Fed specific tone or wording (e.g., `federal funds rate'). To give some examples, the most hawkish phrases are \textit{"Inflation remains too high and requires further policy tightening"} and \textit{"Rising inflationary pressures call for higher interest rates."}. Highly dovish sentences are \textit{"Economic conditions warrant maintaining a low policy rate.} and \textit{"We will continue asset purchases to support favorable financing conditions."}. The hawkish and dovish phrases span topics including inflation, conventional monetary policy, and unconventional measures.
To give some examples for the economic environment seed phrases, a positive environment is captured by phrases such as \textit{"The economy is expanding at a solid pace with strong job growth."}, \textit{"Consumer demand remains strong and supports continued growth."}. Negative sentences comprise \textit{"Weak demand and tight credit conditions are restraining growth."}, or \textit{"Household spending remains constrained by low income growth."}. The full list of phrases are provided in  Tables~\ref{tab:seed_monetary_policy} to \ref{tab:seed_uncertainty} in the appendix. These seed phrases should reflect  the two polar opposites for each dimension and ideally, these dimensions are orthogonal to each other. Hence, in a second step we assess the orthogonality of the embedded seed phrases using PCA and pairwise scatter plots shown in Figures~\ref{fig:seed_scatter} and \ref{fig:seed_pca} in the appendix. This process yielded a final seed set of 11 hawkish and 13 dovish phrases, 14 positive and 13 negative economic outlook phrases, and 15 high-uncertainty and 13 low-uncertainty phrases. 

\begin{figure}[H]
\caption{Distribution of scores for seed phrases}\label{fig:scores_hist}
\centering
\begin{subfigure}[t]{0.31\textwidth}
  \centering
  \caption{Monetary sentiment scores}
  \includegraphics[width=\textwidth]{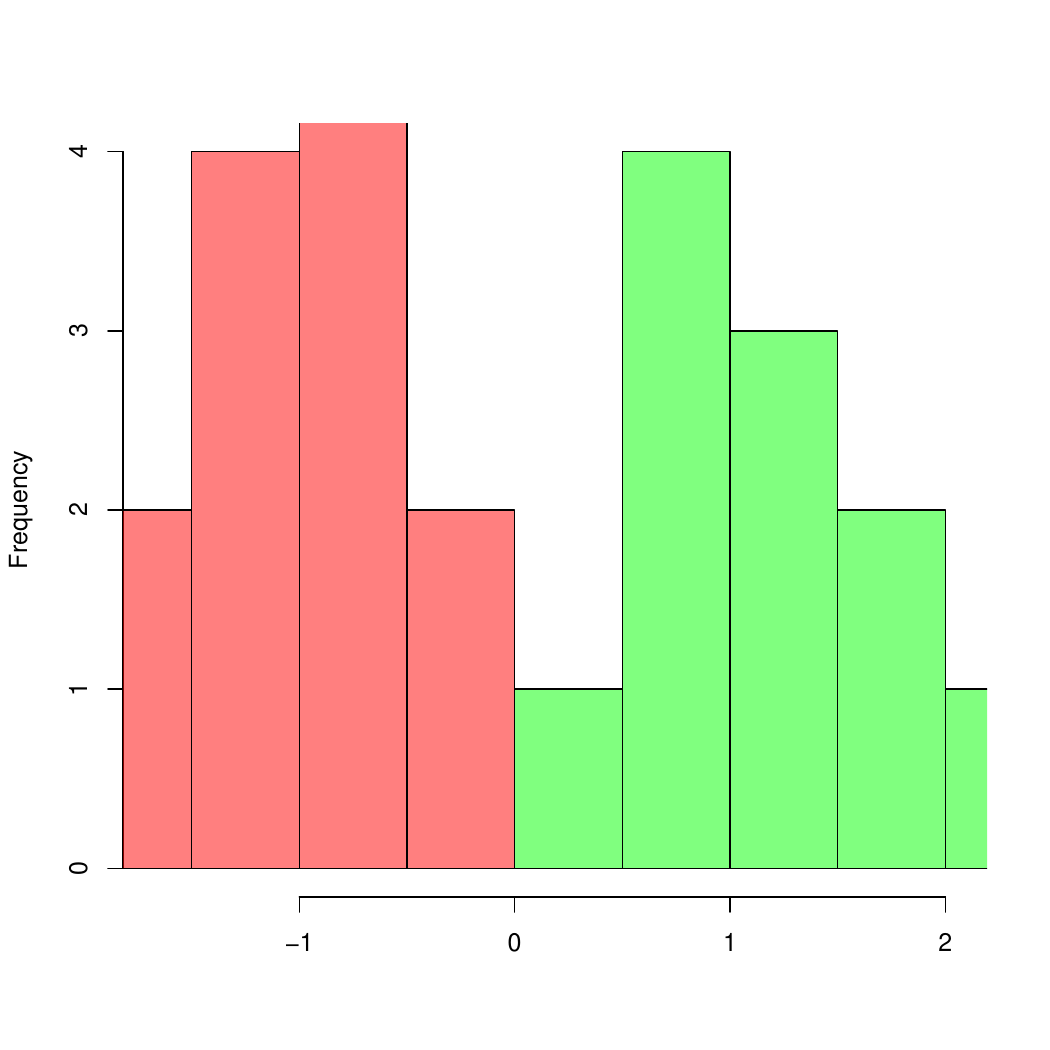}
  \label{fig:mp_scores_hist}
\end{subfigure}
\hfill
\begin{subfigure}[t]{0.31\textwidth}
  \centering
  \caption{Economic sentiment scores}
  \includegraphics[width=\textwidth]{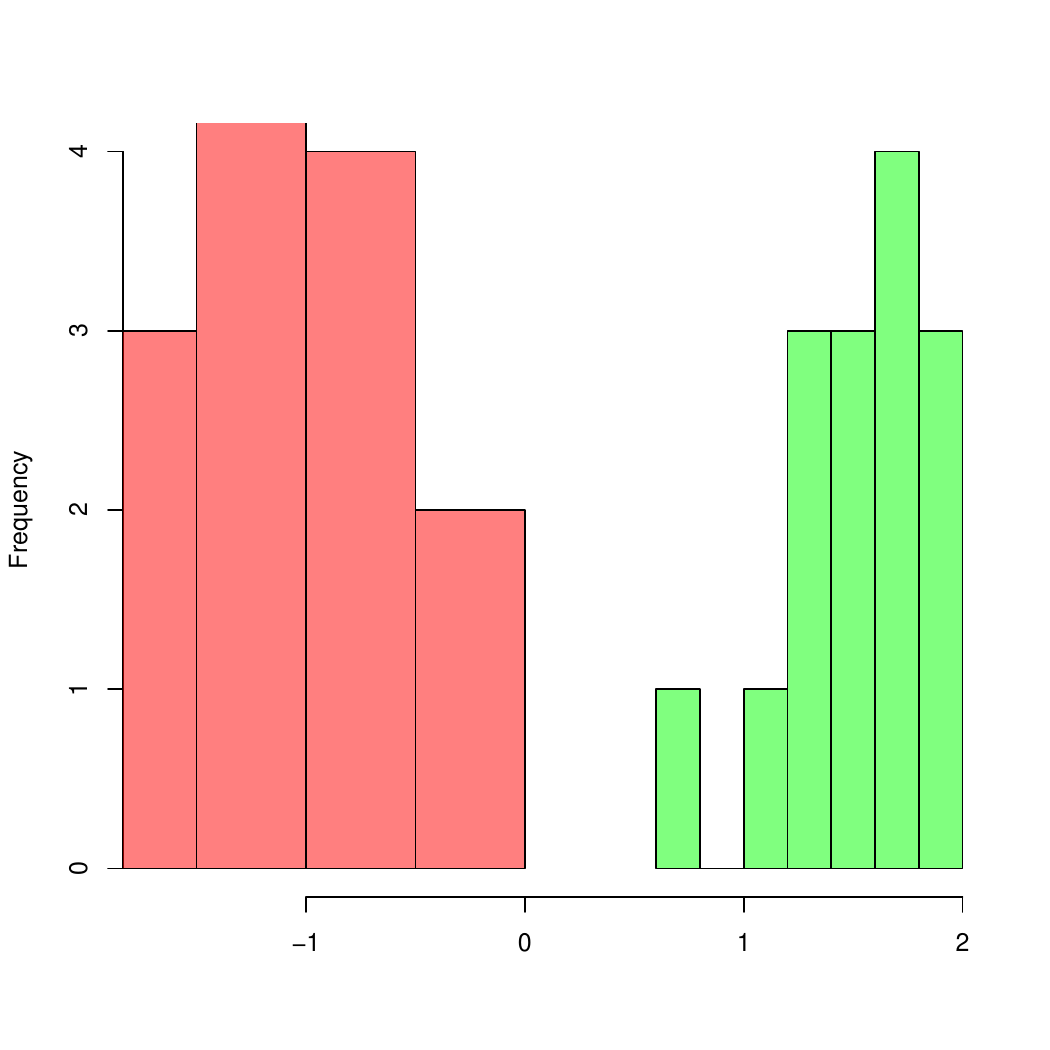}
  \label{fig:econ_scores_hist}
\end{subfigure}
\hfill
\begin{subfigure}[t]{0.31\textwidth}
  \centering
  \caption{Uncertainty scores}
  \includegraphics[width=\textwidth]{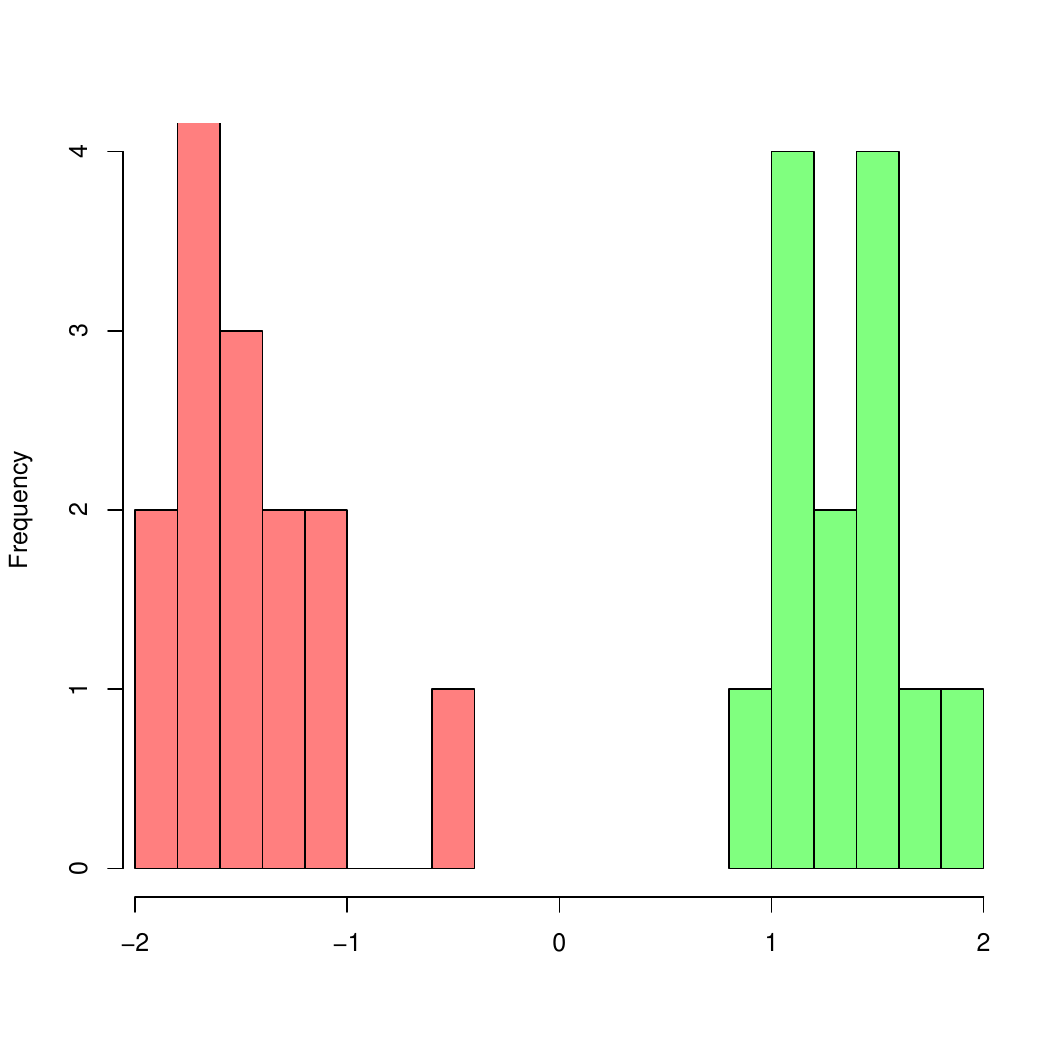}
  \label{fig:unc_scores_hist}
\end{subfigure}
\begin{minipage}{\textwidth}
\footnotesize \textit{Notes}: Distribution of CVP scores for the seed phrases of each dimension. Green bars show scores for the positive pole (hawkish, economic positive, uncertain); red bars show scores for the negative pole (dovish, economic negative, certain). Green and red bars overlapping indicates less separation between poles.
\end{minipage}
\end{figure}

 For this final set of seed phrases, we assess their validity as polar anchors by applying the CVP methodology to the seed phrases themselves. The distribution of the CVP scores for the seed phrases is shown in Figure~\ref{fig:scores_hist} and we would expect a good separation between positive and negative sentiment for all three dimensions in case the seed phrases contain meaningful information.

The plot shows, that for all three dimensions we find a clear separation between positive and negative seed phrases. Two points are worth noting. First, our seed phrases are intentionally generic and institution-neutral, though the approach naturally extends to institution-specific language. Second,  our approach can be easily expanded to capture further sentiment dimensions.

Finally, to derive the sentiment arc, each document must be processed at a chosen level of granularity. This determines how the arc is constructed and governs the degree to which individual segments contribute to the measured sentiment trajectory.

When applying the CVP method, we opt for two strategies.
\begin{enumerate}
    \item A \emph{sentence-level} sentiment arc, where we divide the document into sentences using SpaCy's \citep{spacy} Sentencizer, produce separate embeddings for each sentence, then project them to the concept vectors, and
    \item A \emph{contextual} sentiment arc, where we encode all tokens in the document at once, thereby letting the tokens fully attend to each other, and project token representations directly onto the concept vectors, without pooling. Token scores are then aggregated into equal-width position bins along the document to form the arc. For this purpose, we use the \texttt{nvidia/llama-embed-nemotron-8b} embedding model \citep{nemotron}, which has a large enough context length to fit all introductory statements (32,768 tokens) and, at the time of writing, is the best-performing model on the Massive Multilingual Text Embedding Benchmark \citep{mmteb} when filtering for \textit{Financial}, \textit{Government}, and \textit{Legal} domains. The \texttt{nvidia/llama-embed-nemotron-8b} is a mean-pooling-trained embedding model, which ensures that all tokens belong to the same embedding space.\footnote{Mean-pooling-trained models optimize the sentence representation as the average over all token embeddings, which induces a consistent geometric structure across tokens. CLS-pooled models, by contrast, optimize only the [CLS] token for sentence-level representation, leaving remaining token embeddings without a guarantee of mutual comparability --- a requirement for geometrically meaningful token-level CVP projections.}
\end{enumerate}

We calculate contextual arcs as our benchmark and provide results for sentence-level arcs as a robustness exercise, shown in the appendix.

\section{Sentiment arcs}\label{sec:descriptive}

A sentiment arc is the sequence of CVP scores read from the opening to the close of a statement, tracing how sentiment evolves as the communication unfolds. Because statements differ substantially in length — ECB Introductory Statements average roughly 1,500 words while FOMC statements average roughly 380 words — the position axis is normalized to the interval $[0\%, 100\%]$, mapping the opening token to 0\% and the closing token to 100\% regardless of document length. This normalization allows arcs from different statements to be averaged on a common scale. To reduce token-level noise while preserving within-statement structure, each document-level arc is first smoothed with a LOESS curve (span $= 0.6$, degree 2) evaluated on a 100-point grid, and point-wise means and 95\% confidence intervals are then computed across statements.

\begin{figure}[H]
\caption{Full-sample average sentiment arcs}\label{fig:sent_unconditional}
\centering
\begin{subfigure}[t]{0.48\textwidth}
  \caption{ECB}
  \centering
  \includegraphics[width=\textwidth]{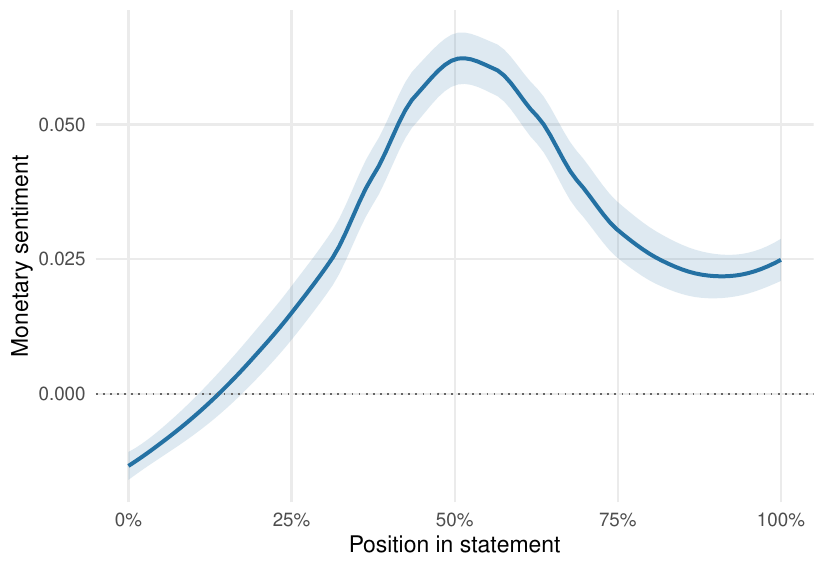}\\[2pt]
  \includegraphics[width=\textwidth]{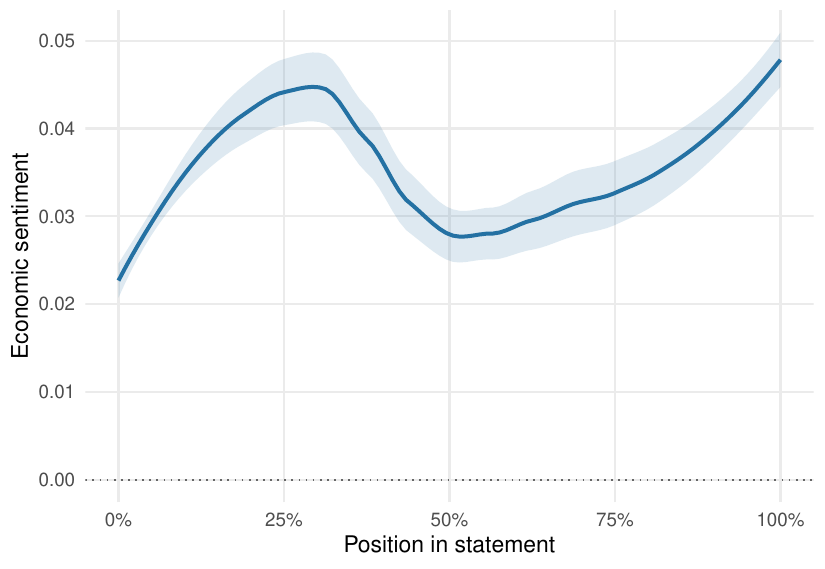}\\[2pt]
  \includegraphics[width=\textwidth]{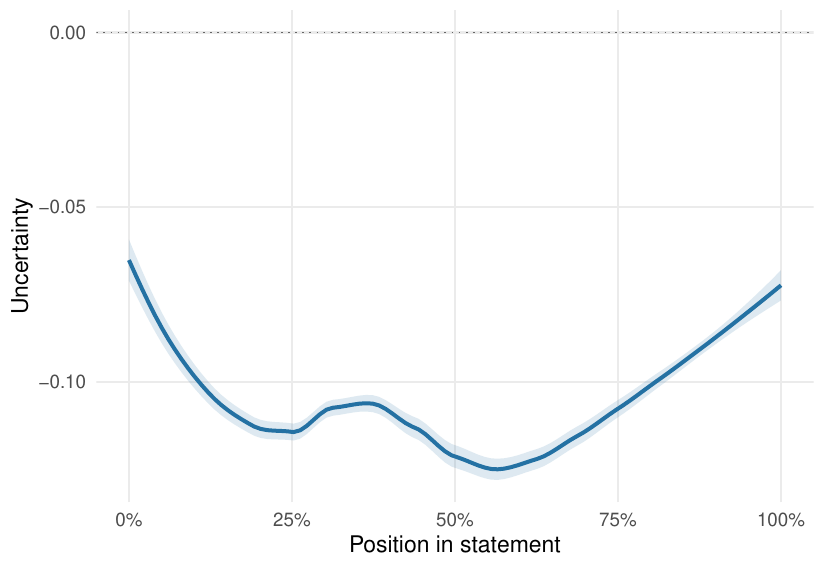}
\end{subfigure}
\hfill
\begin{subfigure}[t]{0.48\textwidth}
  \caption{Fed}
  \centering
  \includegraphics[width=\textwidth]{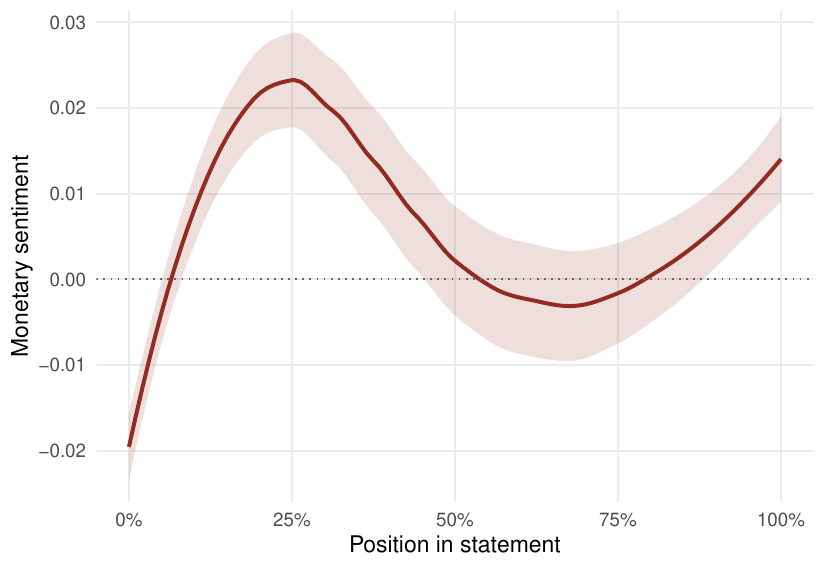}\\[2pt]
  \includegraphics[width=\textwidth]{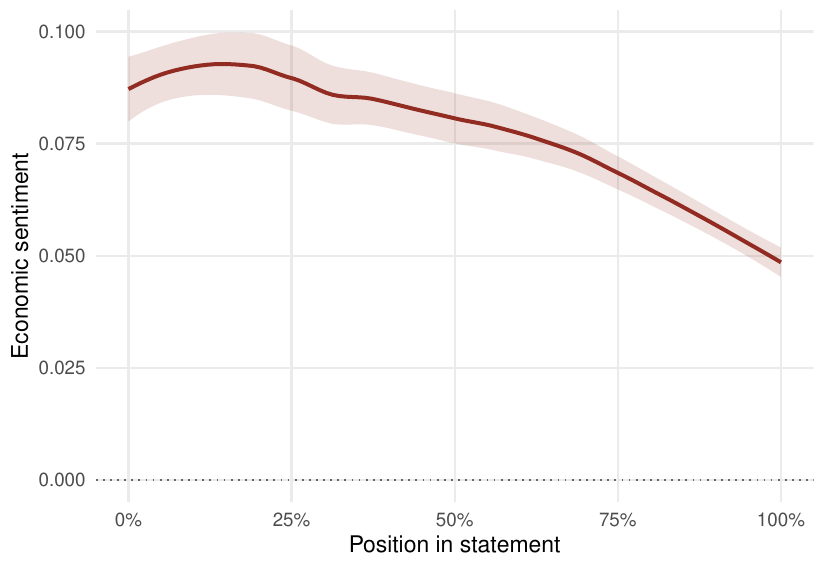}\\[2pt]
  \includegraphics[width=\textwidth]{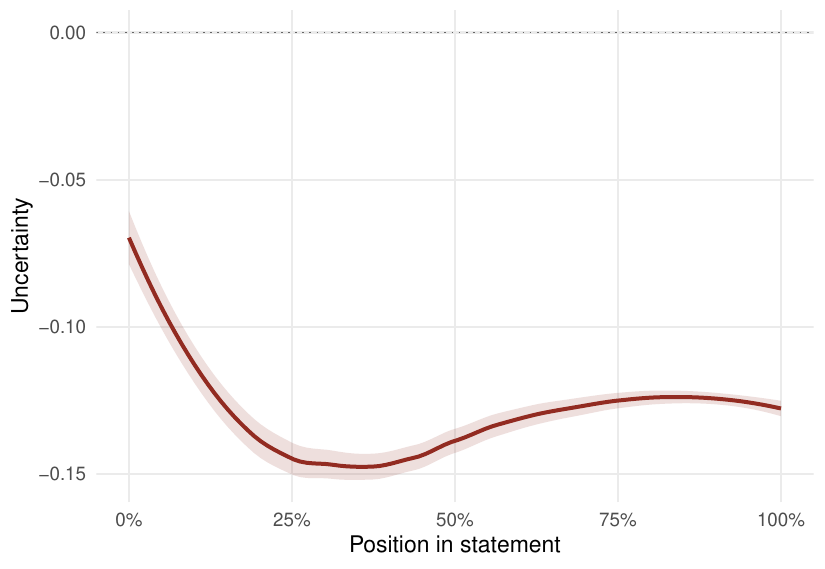}
\end{subfigure}
\begin{minipage}{\textwidth}
\footnotesize \textit{Notes}: LOESS-smoothed (span $= 0.6$) full-sample average sentiment arcs. Each panel shows, from top to bottom: monetary sentiment (hawkish positive), economic sentiment (optimistic positive), and uncertainty sentiment (uncertain positive). The $x$-axis is the normalized position within the statement (0\% = opening, 100\% = closing token), mapping all statements to a common scale regardless of length. Shaded bands are 95\% pointwise confidence intervals across statements. ECB: $N = 269$ statements, 1998--2025. Fed: $N = 219$ statements, 2000--2025. 
\end{minipage}
\end{figure}

Figure~\ref{fig:sent_unconditional} shows the full-sample average arcs for the ECB (left) and the Fed (right), pooled across all statements in the respective corpora. Each panel shows the arcs for the three dimensions analyzed in this study: monetary sentiment (hawkish positive), economic sentiment (optimistic positive), and uncertainty (uncertain positive). The dotted horizontal line marks zero.

Several features of Figure~\ref{fig:sent_unconditional} are worth noting. For the ECB, as the statement opens with the policy decision --- a factual announcement --- all three sentiment measures start close to zero. Monetary sentiment is then on average positive throughout the statement and exhibits some within-statement variation consistent with the two-pillar structure: the early portion, which follows the decision announcement and opens the economic analysis pillar, tends to set the hawkish-dovish tone that the remainder elaborates. Economic sentiment follows a similar within-statement pattern, positive on average and somewhat more front-loaded, with a trajectory that tracks the progression from economic analysis through monetary analysis to the cross-checking conclusion. That said, the average arc described here is a general pattern across all statements; deviations from it --- for instance during crisis periods or ahead of rate changes --- are of particular interest and will be examined more formally in the next section. The uncertainty sentiment arc shows that ECB language is, on average, modestly above zero throughout, reflecting the institution's characteristically deliberate and hedged prose. For the Fed, the shorter and more templated statement structure produces economic sentiment and uncertainty sentiment arcs that are much flatter, with less within-statement variation, consistent with the compressed, formulaic design of FOMC statements. The peak of the economic sentiment indicator falls at the opening of the statement, consistent with the FOMC's structure that leads with the economic environment; the predominantly positive growth conditions the US faced over the sample period are reflected in scores above zero throughout. The only arc that shows considerable within-statement variation is the monetary sentiment one. Here, the peak is reached quickly in the first part of the statement, falls back toward zero, before picking up again in the later portion containing the forward guidance and the factors the Committee will monitor ahead of the next decision.


To verify that the CVP scores capture meaningful content, Table~\ref{tab:sentence_examples} reports the highest- and lowest-scoring sentence in each CVP dimension for each institution, extracted from the full corpora. These are the sentences that sit at the extreme poles of each dimension — the most hawkish, the most dovish, the most optimistic, and so on — as measured by mean token-level CVP projection.

\begin{table}[H]
\centering
\caption{Highest and lowest scoring sentences by CVP dimension}\label{tab:sentence_examples}
\small
\begin{tabular}{llllp{7.8cm}}
\toprule
Dimension & Pole & Date & Score & Sentence \\
\midrule
\multicolumn{5}{l}{\textbf{Panel A: ECB}} \\
\midrule
MP stance   & Hawkish    & 2022-03 & $+0.237$ & Price rises have also become more broadly based. \\[3pt]
MP stance   & Dovish     & 2020-09 & $-0.124$ & We will also continue to provide ample liquidity through our refinancing operations. \\[3pt]
Economic    & Optimistic & 2018-03 & $+0.209$ & The latest economic data and survey results indicate continued strong and broad-based growth momentum. \\[3pt]
Economic    & Pessimistic & 2022-04 & $-0.140$ & Some sectors face growing difficulties in sourcing their inputs, which is disrupting production. \\[3pt]
Uncertainty & Uncertain  & 2002-11 & $+0.085$ & The view has prevailed to keep interest rates unchanged. \\[3pt]
Uncertainty & Certain    & 2010-10 & $-0.259$ & Accordingly, the Governing Council will continue to monitor all developments over the period ahead very closely. \\
\midrule
\multicolumn{5}{l}{\textbf{Panel B: Fed}} \\
\midrule
MP stance   & Hawkish    & 2022-07 & $+0.214$ & The Committee is highly attentive to inflation risks. \\[3pt]
MP stance   & Dovish     & 2015-09 & $-0.111$ & This policy, by keeping the Committee's holdings of longer-term securities at sizable levels, should help maintain accommodative financial conditions. \\[3pt]
Economic    & Optimistic & 2018-08 & $+0.294$ & Household spending and business fixed investment have grown strongly. \\[3pt]
Economic    & Pessimistic & 2020-03 & $-0.132$ & Global financial conditions have also been significantly affected. \\[3pt]
Uncertainty & Uncertain  & 2020-03 & $+0.112$ & The coronavirus outbreak has harmed communities and disrupted economic activity in many countries, including the United States. \\[3pt]
Uncertainty & Certain    & 2006-01 & $-0.263$ & Core inflation has stayed relatively low in recent months and longer-term inflation expectations remain contained. \\
\bottomrule
\end{tabular}
\begin{minipage}{\textwidth}
\footnotesize \textit{Notes}: For each institution and CVP dimension, the table reports the sentence with the highest score (positive pole) and the lowest score (negative pole) across all statements in the corpus. Scores are the mean of token-level CVP projections for tokens belonging to that sentence, computed from contextual arcs. Positive scores reflect the positive pole of each dimension: hawkish for monetary sentiment, optimistic for economic sentiment, and uncertain for the uncertainty sentiment dimension. Sentences with fewer than three tokens or fewer than 40 characters are excluded. The uncertainty dimension scores reflect the degree of epistemic uncertainty in language: the high-scoring sentence scores as highly uncertain while the low-scoring sentence reflects low-uncertainty, assertive language.\end{minipage}
\end{table}

The sentences in Table~\ref{tab:sentence_examples} are readily interpretable. On the monetary sentiment dimension, the highest-scoring ECB sentence — \textit{``Price rises have also become more broadly based''} (March 2022) — captures the broadening of inflationary pressure at the onset of the tightening cycle, while the lowest-scoring one reflects the unconventional liquidity operations of the pandemic period. For the Fed, \textit{``The Committee is highly attentive to inflation risks''} (July 2022) is the single most hawkish sentence in the FOMC corpus, and is indeed the language the Fed introduced when it began its most aggressive tightening cycle in four decades. On the economic sentiment dimension, the top-scoring sentences describe periods of strong, broad-based expansion, while the bottom-scoring sentences characterize supply disruptions and deteriorating financial conditions. The uncertainty sentiment dimension warrants a brief note on interpretation: it scores the degree of epistemic uncertainty in language, not the valence of economic conditions. The highest-scoring sentences (positive scores) reflect language that expresses hedging, qualification, or acknowledged uncertainty; the lowest-scoring sentences (negative scores) tend to use more assertive, settled language. Across all dimensions and both institutions, the extreme sentences align with the historical episodes one would expect, supporting the validity of the CVP scoring approach.

﻿
\section{Predictive regressions}\label{sec:results}

We study whether the \emph{shape} of sentiment arcs — not just their average level — carries predictive information, organized around two conceptually distinct perspectives on how communication design operates. From the \emph{sender's perspective}, the central bank actively structures its narrative: different policy environments call for different arc shapes, and this deliberate structure should be recoverable in the predictive relationship between arc features and subsequent policy decisions. From the \emph{receiver's perspective}, the sequential structure of a statement shapes how audiences process and respond to it: if market participants and households are subject to peak-end effects \citep{Mueller2019}, the arc — not just its average — should affect how beliefs are updated. These two perspectives motivate two families of regressions:

\begin{align}
  \text{Sender (rate changes):}\quad \Delta i_{t+k} &= \alpha + \boldsymbol{\beta}'\,\mathbf{x}^{\text{bench}}_t  + \boldsymbol{\beta}'\,\mathbf{x}^{\text{arc}}_t + \varepsilon_t \label{eq:rate}\\[4pt]
  \text{Receiver (inflation expectations):}\quad \pi^{e}_{t+1} &= \alpha + \boldsymbol{\beta}'\,\mathbf{x}^{\text{bench}}_t +  \boldsymbol{\beta}'\,\mathbf{x}^{\text{arc}}_t + \gamma\,\Delta p^{\mathrm{yoy}}_t + \varepsilon_t \label{eq:infexp}
\end{align}

In equation~(\ref{eq:rate}), $\Delta i_{t+k}$ is the policy rate change decided $k$ meetings ahead of statement $t$, for $k = 1, 2$; both horizons are studied in Section~\ref{sec:rate_pred}. Because meetings are typically held at six-week intervals, this reflects the signaling horizon of each statement rather than a literal one-period-ahead forecast. In equation~(\ref{eq:infexp}), $\pi^{e}_{t+1}$ is the next-month level of professional forecasters' one-year-ahead inflation expectations and $\Delta p^{\mathrm{yoy}}_t$ is the actual year-over-year inflation rate in month $t$, which controls for the persistent level of expectations. Since any press conference in month $t$ necessarily precedes the Consensus Economics survey conducted around the 10th of month $t+1$, timing is unambiguous and no restriction on the day of the press conference is needed.

The feature vectors are constructed as follows. The benchmark $\mathbf{x}^{\text{bench}}$ contains the Picault--Renault dictionary scores (\texttt{mp\_pr}, \texttt{ec\_pr}) described in Section~\ref{sec:data}. The arc features consist of the Nelson--Siegel factors — level, slope, and curvature — estimated separately for each of the three arcs, together with the Hurst exponent for each dimension, capturing arc persistence \citep{Bizzoni2022}. The \emph{level} captures the overall average sentiment across the statement; the \emph{slope} captures the directional trend, that is, whether sentiment rises or falls from the opening to the close of the statement; and the \emph{curvature} captures non-linear within-statement structure, distinguishing statements that build toward a central argument from those that open and close at higher intensity than the analytical middle. We then build a model ladder: M1 uses $\mathbf{x}^{\text{bench}}$ alone; M2 replaces the dictionary scores with CVP document means, providing a natural midpoint between the lexicon baseline and the full arc approach; M3 adds the arc characteristics for the monetary sentiment dimension; M4 extends M3 with arc features of the economic sentiment dimension; the Full arc specification further adds arc features of the uncertainty sentiment dimension. This ladder approach allows us to assess which dimension contributes predictive content. All arc measures are derived from contextual arcs; Section~\ref{sec:arc_comparison} confirms that contextual arcs substantially outperform sentence-level arcs.

\subsection{Arc structure as policy signal}\label{sec:rate_pred}

\begin{figure}[H]
\caption{Sentiment arcs before rate changes}\label{fig:sent_rate}
\centering
\begin{subfigure}[t]{0.48\textwidth}
  \caption{ECB}
  \centering
  \includegraphics[width=\textwidth]{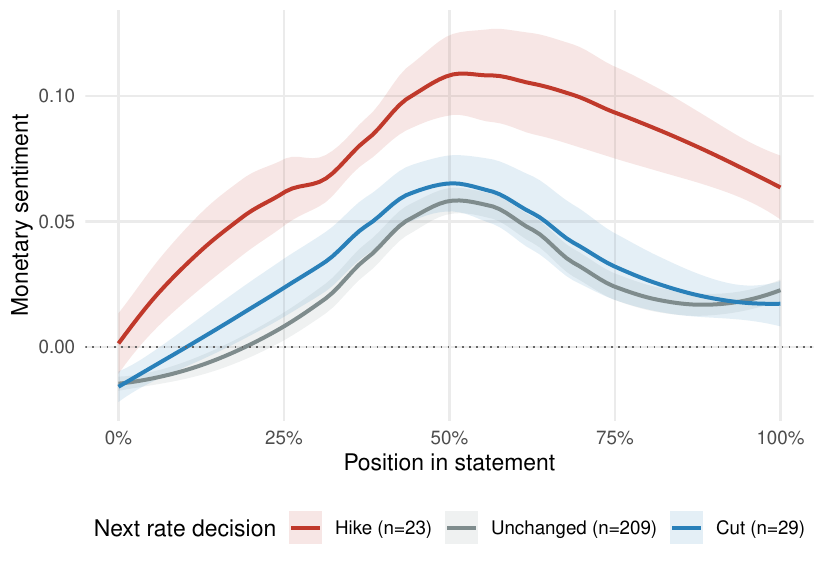}\\[2pt]
  \includegraphics[width=\textwidth]{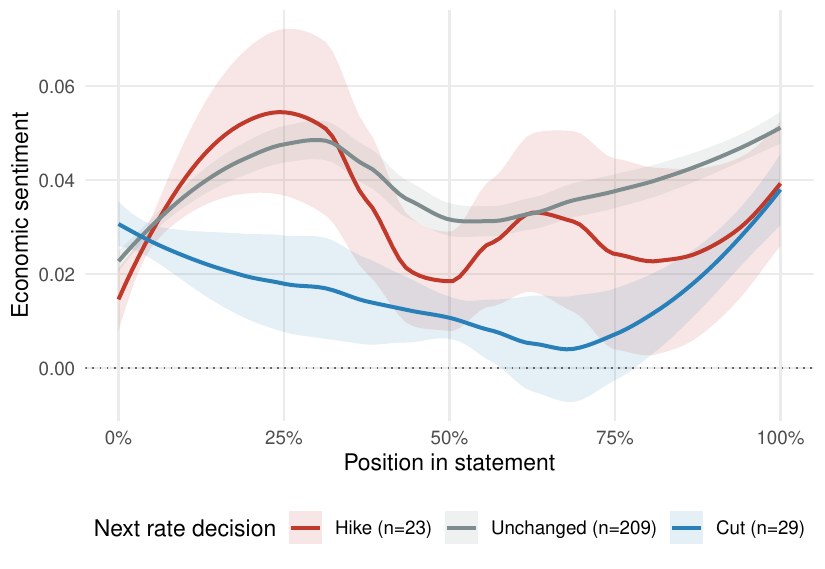}\\[2pt]
  \includegraphics[width=\textwidth]{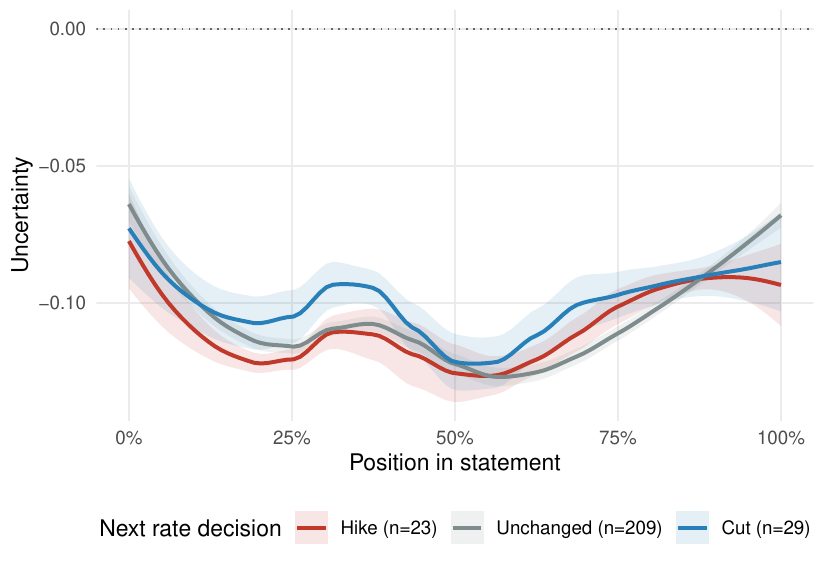}
\end{subfigure}
\hfill
\begin{subfigure}[t]{0.48\textwidth}
  \caption{Fed}
  \centering
  \includegraphics[width=\textwidth]{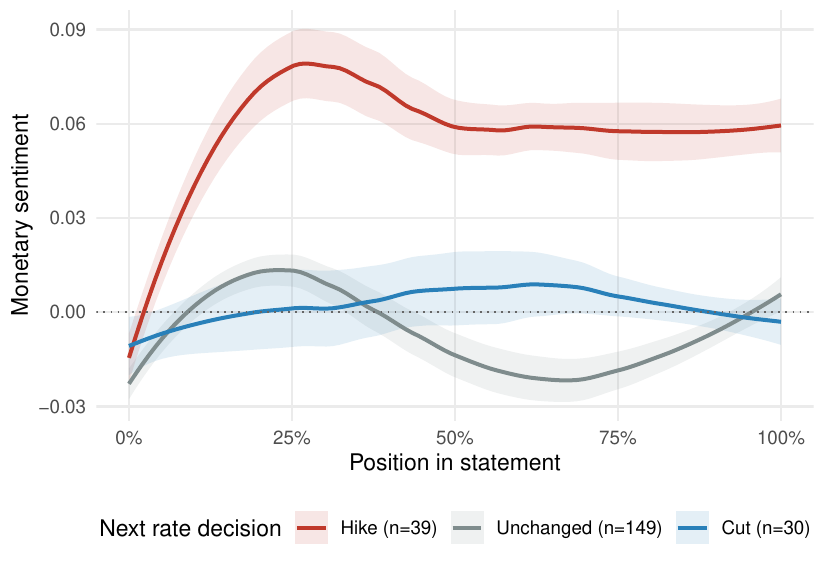}\\[2pt]
  \includegraphics[width=\textwidth]{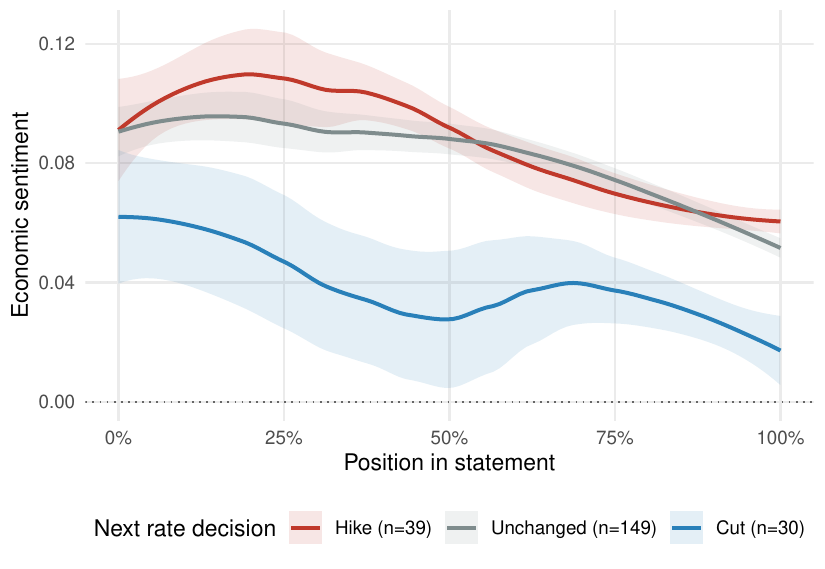}\\[2pt]
  \includegraphics[width=\textwidth]{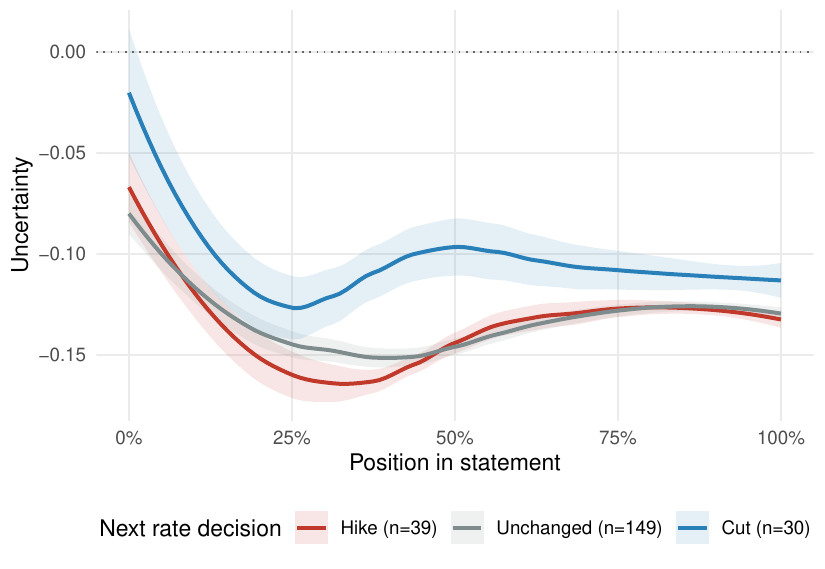}
\end{subfigure}
\begin{minipage}{\textwidth}
\footnotesize \textit{Notes}: LOESS-smoothed (span $= 0.6$) sentiment arcs grouped by the subsequent policy rate decision: hike (red), unchanged (grey), cut (blue). Each panel shows, from top to bottom: monetary sentiment (hawkish positive), economic sentiment (optimistic positive), and uncertainty sentiment (uncertain positive). The $x$-axis is normalized position within statement (0\% = opening, 100\% = close). Shaded bands are 95\% pointwise confidence intervals across statements.\end{minipage}
\end{figure}

Figure~\ref{fig:sent_rate} provides a first visual indication of whether arc shape tracks the subsequent rate decision. Sentiment arcs conditioned on the next meeting's outcome — hike, unchanged, or cut — reveal systematic differences: statements preceding rate hikes display distinctly more hawkish monetary sentiment for both institutions and more positive economic sentiment throughout, with the differences more pronounced for the Fed than for the ECB. In addition and for the Fed, we see an elevated uncertainty arc before rate cuts. These patterns motivate the regression framework below.

Table~\ref{tab:rate_t1_main} presents the model ladder for the rate change regression~(\ref{eq:rate}) for the ECB (left panel) and the Fed (right panel) side by side.

\begin{sidewaystable}[htbp]
\centering
\caption{Rate change regressions: ECB and Fed ($\Delta i_{t+1}$)}\label{tab:rate_t1_main}
\footnotesize
\begin{tabular}{lccccccccccc}
\toprule
 & \multicolumn{5}{c}{\textbf{ECB} ($N = 261$)} & \multicolumn{5}{c}{\textbf{Fed} ($N = 218$)} \\
\cmidrule(lr){2-6}\cmidrule(lr){7-11}
 & M1 & M2 & M3 & M4 & Full arc & M1 & M2 & M3 & M4 & Full arc \\
 & (Bench.) & (Mean) & (MP) & (MP+EC) & (MP+EC+UNC) & (Bench.) & (Mean) & (MP) & (MP+EC) & (MP+EC+UNC) \\
\midrule
(Intercept) & $-0.019$ & $-0.114^{***}$ & $0.819^{***}$ & $0.552^{**}$ & $0.531^{*}$ & $-0.013$ & $-0.229^{***}$ & $0.073$ & $-0.392^{.}$ & $-0.590^{*}$ \\
 & $(-1.343)$ & $(-3.836)$ & $(4.603)$ & $(2.694)$ & $(2.065)$ & $(-0.663)$ & $(-6.615)$ & $(0.481)$ & $(-1.864)$ & $(-2.363)$ \\
\cmidrule(lr){1-6}\cmidrule(lr){7-11}
\texttt{mp\_pr} & $-0.031$ &  &  &  &  & $-0.086^{.}$ &  &  &  &  \\
 & $(-0.965)$ &  &  &  &  & $(-1.721)$ &  &  &  &  \\
\texttt{ec\_pr} & $0.224^{***}$ &  &  &  &  & $0.248^{***}$ &  &  &  &  \\
 & $(3.768)$ &  &  &  &  & $(5.001)$ &  &  &  &  \\
\cmidrule(lr){1-6}\cmidrule(lr){7-11}
\texttt{mp\_mean} &  & $1.462^{***}$ &  &  &  &  & $2.703^{***}$ &  &  &  \\
 &  & $(3.675)$ &  &  &  &  & $(6.475)$ &  &  &  \\
\texttt{econ\_mean} &  & $1.904^{**}$ &  &  &  &  & $2.618^{***}$ &  &  &  \\
 &  & $(3.191)$ &  &  &  &  & $(6.424)$ &  &  &  \\
\cmidrule(lr){1-6}\cmidrule(lr){7-11}
\texttt{mp\_level} &  &  & $2.338^{***}$ & $4.874^{***}$ & $4.908^{***}$ &  &  & $2.729^{***}$ & $1.744^{**}$ & $3.227^{***}$ \\
 &  &  & $(3.700)$ & $(5.591)$ & $(5.266)$ &  &  & $(3.953)$ & $(2.794)$ & $(4.690)$ \\
\texttt{mp\_slope} &  &  & $0.902$ & $3.157^{***}$ & $3.099^{***}$ &  &  & $1.195^{*}$ & $-0.170$ & $0.831$ \\
 &  &  & $(1.455)$ & $(4.117)$ & $(3.697)$ &  &  & $(2.403)$ & $(-0.351)$ & $(1.593)$ \\
\texttt{mp\_curve} &  &  & $0.060$ & $0.449^{**}$ & $0.382^{*}$ &  &  & $0.813^{***}$ & $0.526^{***}$ & $0.629^{***}$ \\
 &  &  & $(0.480)$ & $(3.014)$ & $(2.366)$ &  &  & $(6.331)$ & $(4.254)$ & $(4.975)$ \\
\texttt{hurst\_mp} &  &  & $-1.349^{***}$ & $-1.114^{***}$ & $-1.049^{***}$ &  &  & $-0.219$ & $0.181$ & $0.293$ \\
 &  &  & $(-4.689)$ & $(-3.898)$ & $(-3.439)$ &  &  & $(-0.883)$ & $(0.775)$ & $(1.291)$ \\
\cmidrule(lr){1-6}\cmidrule(lr){7-11}
\texttt{econ\_level} &  &  &  & $5.976^{***}$ & $4.809^{**}$ &  &  &  & $4.986^{***}$ & $3.915^{***}$ \\
 &  &  &  & $(4.082)$ & $(2.795)$ &  &  &  & $(7.498)$ & $(4.705)$ \\
\texttt{econ\_slope} &  &  &  & $3.572^{**}$ & $2.502^{.}$ &  &  &  & $0.441$ & $-0.925^{.}$ \\
 &  &  &  & $(3.298)$ & $(1.894)$ &  &  &  & $(1.499)$ & $(-1.963)$ \\
\texttt{econ\_curve} &  &  &  & $0.920^{***}$ & $0.651^{*}$ &  &  &  & $0.578^{***}$ & $0.525^{***}$ \\
 &  &  &  & $(4.415)$ & $(2.495)$ &  &  &  & $(6.686)$ & $(3.879)$ \\
\texttt{hurst\_econ} &  &  &  & $-0.048$ & $-0.120$ &  &  &  & $-0.124$ & $-0.325$ \\
 &  &  &  & $(-0.231)$ & $(-0.553)$ &  &  &  & $(-0.583)$ & $(-1.534)$ \\
\cmidrule(lr){1-6}\cmidrule(lr){7-11}
\texttt{unc\_level} &  &  &  &  & $-0.881$ &  &  &  &  & $-1.068$ \\
 &  &  &  &  & $(-0.814)$ &  &  &  &  & $(-1.032)$ \\
\texttt{unc\_slope} &  &  &  &  & $-0.653$ &  &  &  &  & $-1.334^{***}$ \\
 &  &  &  &  & $(-1.565)$ &  &  &  &  & $(-3.560)$ \\
\texttt{unc\_curve} &  &  &  &  & $-0.427^{.}$ &  &  &  &  & $0.021$ \\
 &  &  &  &  & $(-1.825)$ &  &  &  &  & $(0.113)$ \\
\texttt{hurst\_unc} &  &  &  &  & $-0.129$ &  &  &  &  & $0.567^{*}$ \\
 &  &  &  &  & $(-0.492)$ &  &  &  &  & $(1.986)$ \\
\midrule
$R^{2}$ & 0.053 & 0.068 & 0.151 & 0.228 & 0.242 & 0.108 & 0.299 & 0.202 & 0.395 & 0.452 \\
Adj.\ $R^{2}$ & 0.045 & 0.060 & 0.138 & 0.204 & 0.205 & 0.099 & 0.292 & 0.187 & 0.372 & 0.420 \\
Residual SE & 0.207 & 0.206 & 0.197 & 0.189 & 0.189 & 0.238 & 0.211 & 0.227 & 0.199 & 0.191 \\
\bottomrule
\end{tabular}
\begin{minipage}{\linewidth}
\footnotesize \textit{Notes}: OLS estimates. $t$-statistics in parentheses. Outcome: one-meeting-ahead policy rate change $\Delta i_{t+1}$. M1 uses Picault--Renault dictionary scores. M2 uses CVP document means. M3 adds Nelson--Siegel factors for the MP arc and \texttt{hurst\_mp}. M4 extends M3 with NS factors for the economic arc and \texttt{hurst\_econ}. Full arc extends M4 with NS factors for the uncertainty sentiment arc and \texttt{hurst\_unc}.Significance: $^{***}p<0.001$, $^{**}p<0.01$, $^{*}p<0.05$, $^{.}p<0.10$.
\end{minipage}
\end{sidewaystable}

For the ECB, the benchmark M1 shows that the Picault--Renault economic sentiment score enters with a positive and highly significant coefficient, while the hawkish--dovish indicator is statistically insignificant, consistent with earlier findings in the literature: lexicon-based hawkishness is a noisy predictor of rate changes, while economic assessments carry cleaner signal. 
M2 turns to the mean of the CVP scores. Fit increases markedly, and both monetary and economic sentiment averages are highly significant, with economic sentiment entering with the larger coefficient. Introducing the arc features of the monetary sentiment dimension delivers the largest single gain in fit (adjusted $R^2 = 0.138$). The level of the arc enters significantly and positively, consistent with more hawkish statements preceding rate increases; slope and curvature are individually insignificant. The Hurst exponent for monetary sentiment enters with a large negative and highly significant coefficient: statements with a more persistent and self-similar monetary sentiment arc — smooth rather than volatile sentiment trajectories — are associated with \emph{lower} subsequent rate changes, pointing to an informative asymmetry between smooth and volatile arc shapes. Model M4 adds arc features of the economic sentiment dimension. Here, the level, slope, and curvature of economic sentiment are all highly significant: a more positive economic assessment (\texttt{econ\_level}), an assessment that becomes progressively more optimistic over the course of the statement (\texttt{econ\_slope}), and one that places greater emphasis on the analytical core (\texttt{econ\_curve}) are each associated with subsequent policy tightening. Adding the economic arc features, also renders the monetary arc slope and curve coefficients significant. Both enter with a positive coefficient. Finally, in the Full arc specification, arc features relating to the uncertainty sentiment dimension all enter negatively, but the coefficients are not precisely estimated, and gains in adjusted $R^2$ are negligible. Arc shape along the monetary sentiment and economic sentiment dimensions thus contributes materially to rate predictions, while arc structure along the uncertainty sentiment dimension does not.

For the Fed, the overall pattern mirrors the ECB but with markedly higher explanatory power across all specifications. The benchmark M1 shows a significant coefficient on \texttt{ec\_pr} and a marginally significant negative coefficient on \texttt{mp\_pr}. CVP mean scores in M2 deliver a substantially larger jump in fit than for the ECB (adjusted $R^2 = 0.292$), suggesting that average CVP tone is a particularly strong signal in Fed communication. M3 yields highly significant level, slope and curvature effects for the monetary sentiment arc. M4 also identifies significant level and curvature contributions from the economic sentiment arc. In the Full arc specification, the uncertainty sentiment arc provides additional predictive content unique to the Fed. The dominant incremental predictor is \texttt{unc\_slope}, which enters with a large negative and highly significant coefficient ($-1.334^{***}$): statements that grow progressively more uncertain toward the close are associated with lower subsequent rate changes, while statements that signal growing confidence toward the close are followed by greater tightening. This is consistent with the Fed resolving its communication into a confident closing stance before acting. The Hurst exponent of the uncertainty sentiment arc also enters positively and significantly ($0.567^{*}$). By contrast, \texttt{hurst\_mp} does not enter significantly for the Fed ($0.293$), unlike the ECB where arc persistence is the key additional feature. This asymmetry reinforces the institutional contrast: for the ECB, \emph{how self-similar the monetary sentiment arc is} encodes the forward rate signal; for the Fed, \emph{how uncertainty sentiment evolves} over the statement is the additional discriminating feature.

Across both institutions, arc features contribute substantially to the share of explained variance. This underscores that the \emph{structure} of economic assessment within a statement — not just its average tone — encodes the forward policy signal.

Table~\ref{tab:rate_t2_main} replicates the model ladder with the two-meeting-ahead rate change $\Delta i_{t+2}$ as the dependent variable. The results mirror those at $t+1$: introducing arc features delivers the largest single gain in fit at both horizons, and arc shape features remain jointly significant. For the ECB, the Hurst exponent of the monetary sentiment arc retains its negative and highly significant coefficient, reinforcing the persistence result across both signaling horizons. For the Fed, the level and curvature of both the monetary sentiment and economic sentiment arcs are robustly significant at both horizons, as well as the slope of the uncertainty sentiment arc and its Hurst exponent.

\begin{sidewaystable}[htbp]
\centering
\caption{Rate change regressions: ECB and Fed ($\Delta i_{t+2}$)}\label{tab:rate_t2_main}
\footnotesize
\begin{tabular}{lccccccccccc}
\toprule
 & \multicolumn{5}{c}{\textbf{ECB} ($N = 260$)} & \multicolumn{5}{c}{\textbf{Fed} ($N = 217$)} \\
\cmidrule(lr){2-6}\cmidrule(lr){7-11}
 & M1 & M2 & M3 & M4 & Full arc & M1 & M2 & M3 & M4 & Full arc \\
 & (Bench.) & (Mean) & (MP) & (MP+EC) & (MP+EC+UNC) & (Bench.) & (Mean) & (MP) & (MP+EC) & (MP+EC+UNC) \\
\midrule
(Intercept) & $-0.037$ & $-0.219^{***}$ & $1.400^{***}$ & $0.942^{**}$ & $0.999^{*}$ & $-0.039$ & $-0.441^{***}$ & $0.151$ & $-0.800^{*}$ & $-1.031^{*}$ \\
 & $(-1.583)$ & $(-4.553)$ & $(4.829)$ & $(2.843)$ & $(2.398)$ & $(-1.116)$ & $(-7.321)$ & $(0.568)$ & $(-2.180)$ & $(-2.348)$ \\
\cmidrule(lr){1-6}\cmidrule(lr){7-11}
\texttt{mp\_pr} & $-0.070$ &  &  &  &  & $-0.222^{*}$ &  &  &  &  \\
 & $(-1.344)$ &  &  &  &  & $(-2.587)$ &  &  &  &  \\
\texttt{ec\_pr} & $0.410^{***}$ &  &  &  &  & $0.528^{***}$ &  &  &  &  \\
 & $(4.207)$ &  &  &  &  & $(6.185)$ &  &  &  &  \\
\cmidrule(lr){1-6}\cmidrule(lr){7-11}
\texttt{mp\_mean} &  & $3.092^{***}$ &  &  &  &  & $4.838^{***}$ &  &  &  \\
 &  & $(4.808)$ &  &  &  &  & $(6.666)$ &  &  &  \\
\texttt{econ\_mean} &  & $3.461^{***}$ &  &  &  &  & $5.051^{***}$ &  &  &  \\
 &  & $(3.572)$ &  &  &  &  & $(7.128)$ &  &  &  \\
\cmidrule(lr){1-6}\cmidrule(lr){7-11}
\texttt{mp\_level} &  &  & $4.086^{***}$ & $8.609^{***}$ & $8.972^{***}$ &  &  & $5.197^{***}$ & $3.523^{**}$ & $6.063^{***}$ \\
 &  &  & $(3.975)$ & $(6.098)$ & $(5.953)$ &  &  & $(4.250)$ & $(3.201)$ & $(4.985)$ \\
\texttt{mp\_slope} &  &  & $1.354$ & $5.258^{***}$ & $5.354^{***}$ &  &  & $2.543^{**}$ & $0.241$ & $1.770^{.}$ \\
 &  &  & $(1.342)$ & $(4.236)$ & $(3.950)$ &  &  & $(2.874)$ & $(0.282)$ & $(1.917)$ \\
\texttt{mp\_curve} &  &  & $0.231$ & $0.937^{***}$ & $0.897^{***}$ &  &  & $1.548^{***}$ & $1.099^{***}$ & $1.258^{***}$ \\
 &  &  & $(1.132)$ & $(3.885)$ & $(3.436)$ &  &  & $(6.861)$ & $(5.087)$ & $(5.664)$ \\
\texttt{hurst\_mp} &  &  & $-2.345^{***}$ & $-2.050^{***}$ & $-1.779^{***}$ &  &  & $-0.435$ & $0.319$ & $0.555$ \\
 &  &  & $(-5.005)$ & $(-4.425)$ & $(-3.603)$ &  &  & $(-0.998)$ & $(0.785)$ & $(1.402)$ \\
\cmidrule(lr){1-6}\cmidrule(lr){7-11}
\texttt{econ\_level} &  &  &  & $9.858^{***}$ & $9.150^{**}$ &  &  &  & $8.463^{***}$ & $7.485^{***}$ \\
 &  &  &  & $(4.148)$ & $(3.281)$ &  &  &  & $(7.288)$ & $(5.157)$ \\
\texttt{econ\_slope} &  &  &  & $5.200^{**}$ & $4.134^{.}$ &  &  &  & $0.601$ & $-1.535^{.}$ \\
 &  &  &  & $(2.965)$ & $(1.933)$ &  &  &  & $(1.171)$ & $(-1.864)$ \\
\texttt{econ\_curve} &  &  &  & $1.359^{***}$ & $1.124^{**}$ &  &  &  & $1.065^{***}$ & $1.137^{***}$ \\
 &  &  &  & $(4.026)$ & $(2.662)$ &  &  &  & $(7.052)$ & $(4.820)$ \\
\texttt{hurst\_econ} &  &  &  & $0.049$ & $0.025$ &  &  &  & $-0.038$ & $-0.394$ \\
 &  &  &  & $(0.147)$ & $(0.072)$ &  &  &  & $(-0.101)$ & $(-1.064)$ \\
\cmidrule(lr){1-6}\cmidrule(lr){7-11}
\texttt{unc\_level} &  &  &  &  & $0.443$ &  &  &  &  & $0.060$ \\
 &  &  &  &  & $(0.254)$ &  &  &  &  & $(0.033)$ \\
\texttt{unc\_slope} &  &  &  &  & $-0.552$ &  &  &  &  & $-1.950^{**}$ \\
 &  &  &  &  & $(-0.819)$ &  &  &  &  & $(-2.984)$ \\
\texttt{unc\_curve} &  &  &  &  & $-0.451$ &  &  &  &  & $0.357$ \\
 &  &  &  &  & $(-1.193)$ &  &  &  &  & $(1.076)$ \\
\texttt{hurst\_unc} &  &  &  &  & $-0.402$ &  &  &  &  & $1.069^{*}$ \\
 &  &  &  &  & $(-0.938)$ &  &  &  &  & $(2.089)$ \\
\midrule
$R^{2}$ & 0.064 & 0.101 & 0.172 & 0.254 & 0.269 & 0.161 & 0.329 & 0.220 & 0.417 & 0.473 \\
Adj.\ $R^{2}$ & 0.057 & 0.094 & 0.159 & 0.231 & 0.234 & 0.153 & 0.323 & 0.206 & 0.395 & 0.442 \\
Residual SE & 0.339 & 0.333 & 0.320 & 0.306 & 0.306 & 0.411 & 0.367 & 0.398 & 0.347 & 0.334 \\
\bottomrule
\end{tabular}
\begin{minipage}{\linewidth}
\footnotesize \textit{Notes}: OLS estimates. $t$-statistics in parentheses. Outcome: two-meeting-ahead policy rate change $\Delta i_{t+2}$. M1 uses Picault--Renault dictionary scores. M2 uses CVP document means. M3 adds Nelson--Siegel factors for the MP arc and \texttt{hurst\_mp}. M4 extends M3 with NS factors for the economic arc and \texttt{hurst\_econ}. Full arc extends M4 with NS factors for the uncertainty sentiment arc and \texttt{hurst\_unc}.Significance: $^{***}p<0.001$, $^{**}p<0.01$, $^{*}p<0.05$, $^{.}p<0.10$.
\end{minipage}
\end{sidewaystable}

\subsubsection{Communication discipline under stress}\label{sec:crisis}

A further question is whether the arc--rate relationship reflects stable communication design or whether it breaks down under macroeconomic stress. More specifically, we define crisis periods using established business cycle dating. For the ECB, dates follow the Euro Area Business Cycle Network (EABC) dating committee: the Global Financial Crisis (April 2008--March 2009), the euro area sovereign debt crisis   (July 2011--March 2013), and the COVID crisis (January--June 2020). For the Fed, the euro area sovereign debt crisis is      excluded as it did not constitute a systemic stress episode for the United States; the two remaining crises are dated   following the NBER Business Cycle Dating Committee: the Global Financial Crisis (December 2007--June 2009) and the COVID  recession (February--April 2020).

\begin{figure}[H]
\caption{Sentiment arcs during crisis and non-crisis periods}\label{fig:sent_crisis}
\centering
\begin{subfigure}[t]{0.48\textwidth}
  \caption{ECB}
  \centering
  \includegraphics[width=\textwidth]{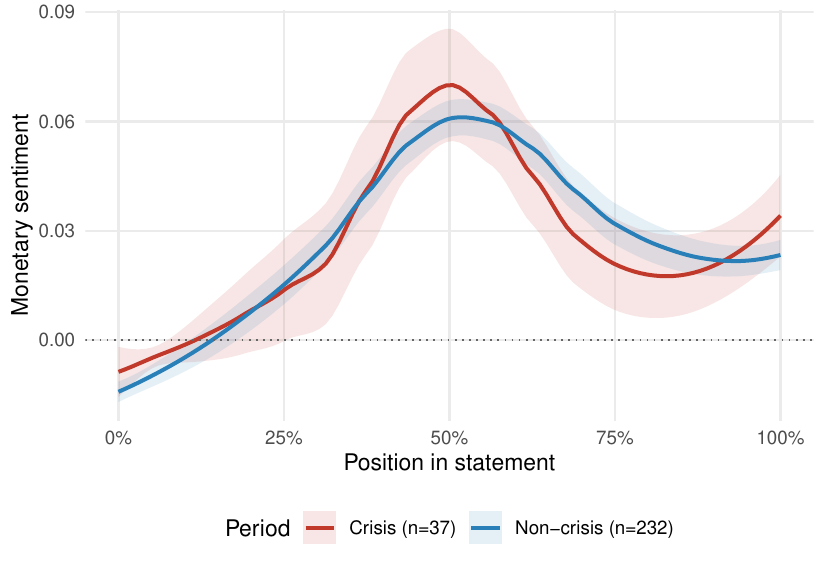}\\[2pt]
  \includegraphics[width=\textwidth]{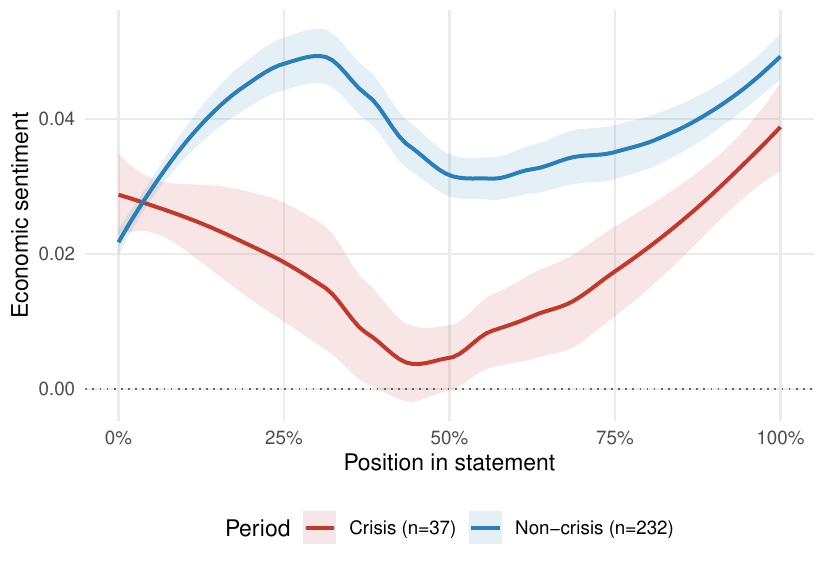}\\[2pt]
  \includegraphics[width=\textwidth]{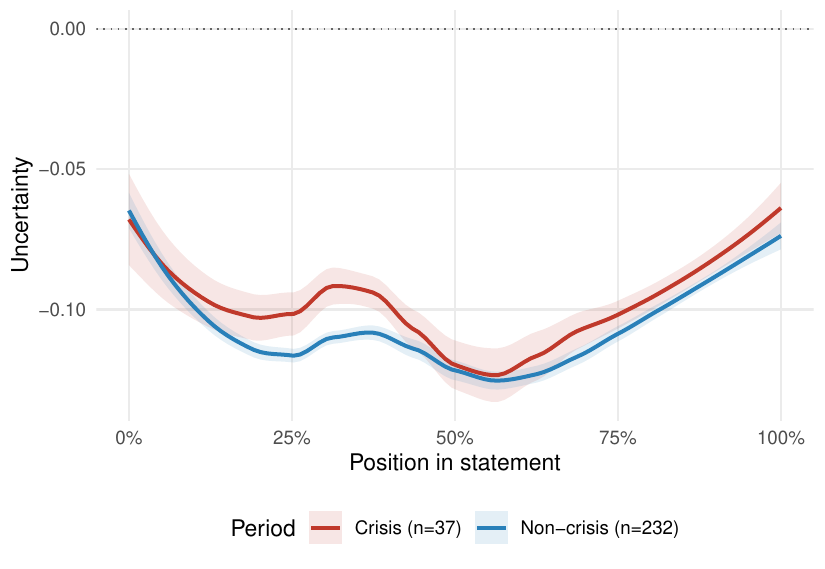}
\end{subfigure}
\hfill
\begin{subfigure}[t]{0.48\textwidth}
  \caption{Fed}
  \centering
  \includegraphics[width=\textwidth]{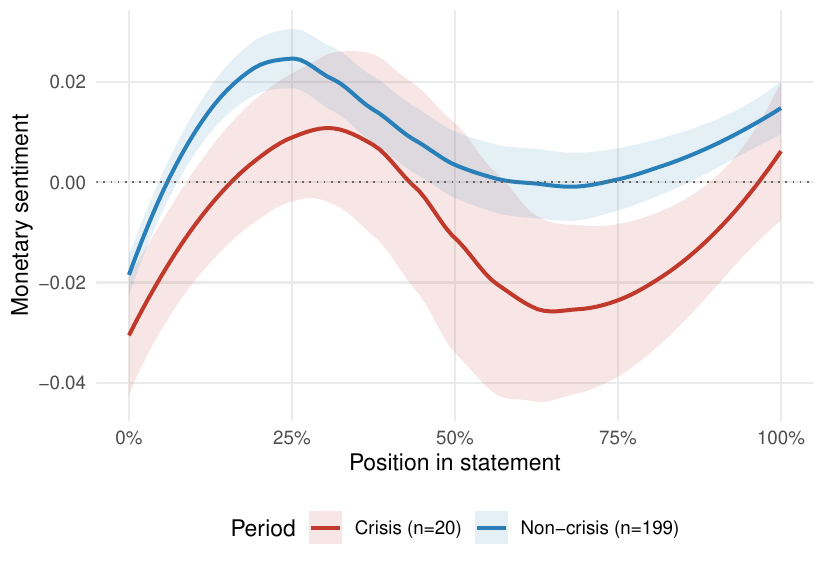}\\[2pt]
  \includegraphics[width=\textwidth]{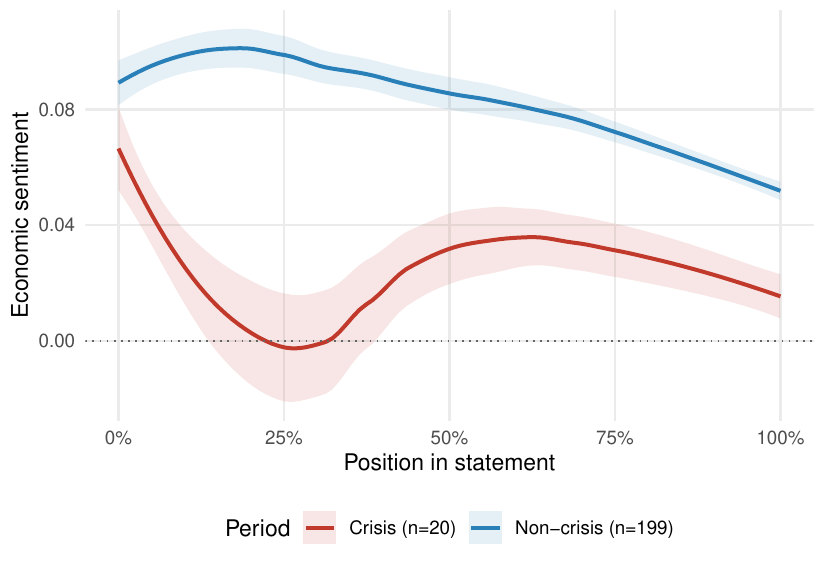}\\[2pt]
  \includegraphics[width=\textwidth]{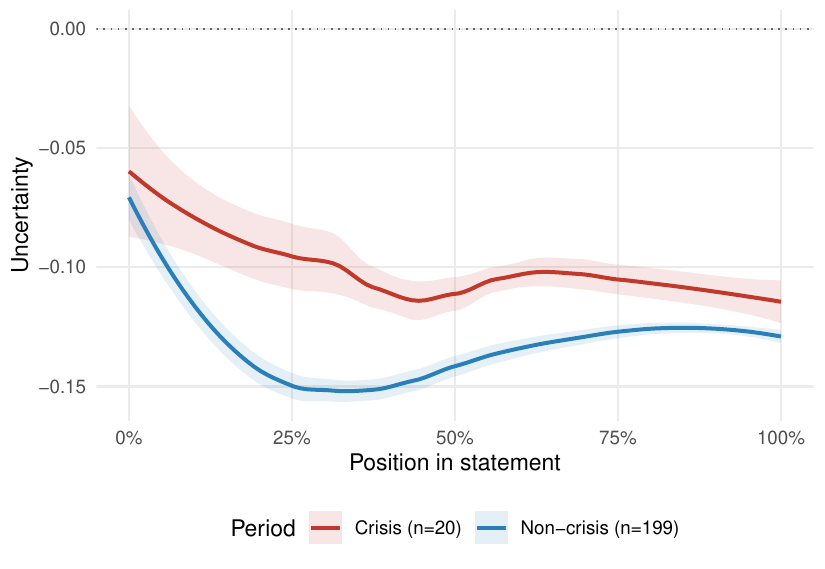}
\end{subfigure}
\begin{minipage}{\textwidth}
\footnotesize \textit{Notes}: LOESS-smoothed (span $= 0.6$) sentiment arcs during crisis (red) and non-crisis (blue) periods. Each panel shows, from top to bottom: monetary sentiment (hawkish positive), economic sentiment (optimistic positive), and uncertainty sentiment (uncertain positive). The $x$-axis is normalized position within statement (0\% = opening, 100\% = close). Shaded bands are 95\% pointwise confidence intervals. ECB crisis periods follow the EABC dating committee: Global Financial Crisis (April 2008--March 2009), euro area sovereign debt crisis (July 2011--March 2013), and COVID crisis (January--June 2020). Fed crisis periods follow the NBER dating committee: Global Financial Crisis (December 2007--June 2009) and COVID recession (February--April 2020); the euro area sovereign debt crisis is excluded.\end{minipage}
\end{figure}

Figure~\ref{fig:sent_crisis} provides visual motivation: sentiment arcs during crisis and non-crisis periods differ systematically in shape, but not along all dimensions and among institutions to the same extent. More specifically, for the ECB monetary sentiment and the uncertainty arcs do not display significant differences between crisis and non-crisis periods. Only the economic sentiment is much more negative during crisis times on average, but shooting up at the end of the statement, possibly to make use of the peak-end bias and to send a positive message. Also the Fed arcs show a more negative economic sentiment arc during crisis periods. However, the Fed also shows a more dovish arc during crisis periods as well as more certainty. The Full arc$\times$C specification augments the full arc model with a crisis dummy and its full set of interactions with arc features of all three dimensions, formalizing these patterns. Table~\ref{tab:rate_crisis} presents results for both the ECB and the Fed.

\begin{table}[H]
\centering
\caption{Rate change regressions: crisis interactions}\label{tab:rate_crisis}
\footnotesize
\begin{tabular}{lcccc}
\toprule
 & \multicolumn{2}{c}{\textbf{ECB} ($N = 261$)} & \multicolumn{2}{c}{\textbf{Fed} ($N = 218$)} \\
\cmidrule(lr){2-3}\cmidrule(lr){4-5}
 & Full arc & Full arc$\times$C & Full arc & Full arc$\times$C \\
\midrule
(Intercept) & $0.531^{*}$ & $0.588^{*}$ & $-0.590^{*}$ & $-0.187$ \\
 & $(2.065)$ & $(2.260)$ & $(-2.363)$ & $(-0.819)$ \\
\cmidrule(lr){1-3}\cmidrule(lr){4-5}
\texttt{mp\_level} & $4.908^{***}$ & $4.625^{***}$ & $3.227^{***}$ & $4.036^{***}$ \\
 & $(5.266)$ & $(4.876)$ & $(4.690)$ & $(6.563)$ \\
\texttt{mp\_slope} & $3.099^{***}$ & $2.495^{**}$ & $0.831$ & $1.795^{***}$ \\
 & $(3.697)$ & $(2.871)$ & $(1.593)$ & $(3.832)$ \\
\texttt{mp\_curve} & $0.382^{*}$ & $0.416^{*}$ & $0.629^{***}$ & $0.827^{***}$ \\
 & $(2.366)$ & $(2.447)$ & $(4.975)$ & $(7.405)$ \\
\texttt{hurst\_mp} & $-1.049^{***}$ & $-1.136^{***}$ & $0.293$ & $0.036$ \\
 & $(-3.439)$ & $(-3.718)$ & $(1.291)$ & $(0.176)$ \\
\cmidrule(lr){1-3}\cmidrule(lr){4-5}
\texttt{econ\_level} & $4.809^{**}$ & $5.328^{**}$ & $3.915^{***}$ & $1.691^{*}$ \\
 & $(2.795)$ & $(2.945)$ & $(4.705)$ & $(2.186)$ \\
\texttt{econ\_slope} & $2.502^{.}$ & $3.224^{*}$ & $-0.925^{.}$ & $-1.382^{**}$ \\
 & $(1.894)$ & $(2.325)$ & $(-1.963)$ & $(-3.243)$ \\
\texttt{econ\_curve} & $0.651^{*}$ & $0.569^{*}$ & $0.525^{***}$ & $0.254^{.}$ \\
 & $(2.495)$ & $(2.102)$ & $(3.879)$ & $(1.920)$ \\
\texttt{hurst\_econ} & $-0.120$ & $-0.032$ & $-0.325$ & $-0.317$ \\
 & $(-0.553)$ & $(-0.146)$ & $(-1.534)$ & $(-1.647)$ \\
\cmidrule(lr){1-3}\cmidrule(lr){4-5}
\texttt{unc\_level} & $-0.881$ & $-0.213$ & $-1.068$ & $-2.019^{*}$ \\
 & $(-0.814)$ & $(-0.184)$ & $(-1.032)$ & $(-2.154)$ \\
\texttt{unc\_slope} & $-0.653$ & $-0.234$ & $-1.334^{***}$ & $-1.442^{***}$ \\
 & $(-1.565)$ & $(-0.518)$ & $(-3.560)$ & $(-4.197)$ \\
\texttt{unc\_curve} & $-0.427^{.}$ & $-0.215$ & $0.021$ & $-0.132$ \\
 & $(-1.825)$ & $(-0.840)$ & $(0.113)$ & $(-0.772)$ \\
\texttt{hurst\_unc} & $-0.129$ & $-0.141$ & $0.567^{*}$ & $0.188$ \\
 & $(-0.492)$ & $(-0.532)$ & $(1.986)$ & $(0.717)$ \\
\cmidrule(lr){1-3}\cmidrule(lr){4-5}
\texttt{crisis} &  & $-0.229$ &  & $-1.300^{*}$ \\
 &  & $(-0.799)$ &  & $(-2.387)$ \\
\texttt{mp\_level} $\times$ crisis &  & $3.244$ &  & $-14.708^{**}$ \\
 &  & $(0.470)$ &  & $(-3.071)$ \\
\texttt{mp\_slope} $\times$ crisis &  & $5.013$ &  & $-15.311^{***}$ \\
 &  & $(0.885)$ &  & $(-4.550)$ \\
\texttt{mp\_curve} $\times$ crisis &  & $0.659$ &  & $-1.112^{.}$ \\
 &  & $(0.661)$ &  & $(-1.853)$ \\
\texttt{econ\_level} $\times$ crisis &  & $-4.784$ &  & $12.665^{*}$ \\
 &  & $(-0.627)$ &  & $(2.454)$ \\
\texttt{econ\_slope} $\times$ crisis &  & $-3.127$ &  & $-4.870^{*}$ \\
 &  & $(-0.599)$ &  & $(-2.036)$ \\
\texttt{econ\_curve} $\times$ crisis &  & $0.259$ &  & $1.270^{*}$ \\
 &  & $(0.261)$ &  & $(2.378)$ \\
\texttt{unc\_level} $\times$ crisis &  & $-5.274$ &  & $-3.014$ \\
 &  & $(-1.391)$ &  & $(-0.607)$ \\
\texttt{unc\_slope} $\times$ crisis &  & $-1.133$ &  & $0.573$ \\
 &  & $(-0.768)$ &  & $(0.551)$ \\
\texttt{unc\_curve} $\times$ crisis &  & $-0.975$ &  & $-1.133$ \\
 &  & $(-1.174)$ &  & $(-1.474)$ \\
\midrule
$R^{2}$ & 0.242 & 0.294 & 0.452 & 0.640 \\
Adj.\ $R^{2}$ & 0.205 & 0.229 & 0.420 & 0.599 \\
Residual SE & 0.189 & 0.186 & 0.191 & 0.159 \\
\bottomrule
\end{tabular}
\begin{minipage}{\textwidth}
\footnotesize \textit{Notes}: OLS estimates. $t$-statistics in parentheses. Outcome: one-meeting-ahead policy rate change $\Delta i_{t+1}$. Full arc model includes Nelson--Siegel factors for all three arc dimensions (monetary sentiment, economic sentiment, uncertainty sentiment) and their Hurst exponents. ECB crisis periods follow the EABC dating committee: Global Financial Crisis (April 2008--March 2009), euro area sovereign debt crisis (July 2011--March 2013), COVID crisis (January--June 2020). Fed crisis periods follow the NBER dating committee: Global Financial Crisis (December 2007--June 2009) and COVID recession (February--April 2020); the euro area sovereign debt crisis is excluded. Significance: $^{***}p<0.001$, $^{**}p<0.01$, $^{*}p<0.05$, $^{.}p<0.10$.
\end{minipage}
\end{table}

The results reveal a striking institutional contrast. For the ECB, the crisis dummy itself is insignificant, and  all interaction terms are statistically insignificant. Overall, the ECB's arc--rate relationship is remarkably stable across crisis and non-crisis periods, consistent with the institution's reputation for rules-based, systematic communication. For the Fed, by contrast, the Full arc$\times$C specification uncovers strong regime heterogeneity. The crisis dummy itself enters with a large negative and statistically significant coefficient. Moreover, the level and slope of the monetary sentiment arc interact significantly and negatively with the crisis dummy: the usual positive relationship between hawkish arc shape and rate increases reverses sharply during crises, implying that even a hawkish-sounding arc does not signal rate hikes when the Fed is responding to an acute shock. The economic sentiment arc interactions are also partially significant: higher economic level in crisis periods is associated with larger rate cuts (positive interaction), consistent with the Fed providing optimistic forward guidance even as it eases aggressively. Finally, the curvature of the uncertainty sentiment arc interacts significantly and negatively with the crisis dummy, implying that statements in which expressed uncertainty peaks in the analytical core — even as the surrounding framing and forward guidance remain more measured — are followed by the largest rate cuts during crises. This pattern is consistent with a communication design in which the central bank explicitly acknowledges the highest uncertainty about the shock's trajectory in the analytical middle of the statement, accompanying the most decisive policy responses.

\subsection{Arc structure and audience responses}\label{sec:audience}

We now turn to the receiver side: does the arc structure of a statement affect how audiences process and respond to it? We focus on professional forecasters updating inflation expectations. If the arc carries purely internal policy signals, its predictive content should be exhausted by the rate regressions above; finding effects on audience responses would instead suggest that the sequential narrative structure shapes perception beyond the policy content itself. As a further extension, we provide results on asset price reactions over the ECB press conference window in Appendix~\ref{sec:appendix_asset}.

\subsubsection{Inflation expectations}\label{sec:infexp}

We examine whether arc features predict professional inflation expectations and forecaster disagreement. \citet{Hubert2017} establishes that ECB qualitative communication and macroeconomic projections jointly affect private inflation expectations, with qualitative communication mattering most at shorter horizons; we ask whether arc structure carries additional predictive content beyond the aggregate tone. The outcome is the next-month Consensus Economics one-year-ahead inflation expectation  $\pi^{e}_{t+1}$, with actual year-over-year inflation $\Delta p^{\mathrm{yoy}}_t$ included as a control to absorb the persistent level component. Consensus Economics provides expectations from professional forecasters such as commercial banks, research institutions or think tanks. Hence, it is more likely that analysts working for these forecasting institutions actually listen to the central banks statements and the arc shape potentially matters. This would probably be different when analyzing household expectations.\footnote{On household attention to central bank communication, see for example \citet{Hajdini2026} using a novel survey or more indirectly \citet{feldkircher2025central} linking Google search data to central bank speech events.} In case there are multiple statements per month, we take the latest one. This yields $N = 225$ matched observations for the ECB and $N = 184$ for the Fed. Table~\ref{tab:infexp} reports results for both institutions.

\begin{table}[H]
\centering
\caption{Inflation expectations and forecaster disagreement regressions}\label{tab:infexp}
\footnotesize
\resizebox{\linewidth}{!}{%
\begin{tabular}{lcccccccc}
\toprule
 & \multicolumn{4}{c}{\textbf{Inflation expectations} ($\pi^{e}_{t+1}$)} & \multicolumn{4}{c}{\textbf{Forecaster disagreement} ($\sigma_{t+1}(\pi^{e})$)} \\
\cmidrule(lr){2-5}\cmidrule(lr){6-9}
 & \multicolumn{2}{c}{ECB ($N=225$)} & \multicolumn{2}{c}{Fed ($N=184$)} & \multicolumn{2}{c}{ECB ($N=225$)} & \multicolumn{2}{c}{Fed ($N=184$)} \\
\cmidrule(lr){2-3}\cmidrule(lr){4-5}\cmidrule(lr){6-7}\cmidrule(lr){8-9}
 & M1 & Full arc & M1 & Full arc & M1 & Full arc & M1 & Full arc \\
\midrule
(Intercept) & $0.694^{***}$ & $0.928^{.}$ & $0.916^{***}$ & $0.829$ & $0.213^{***}$ & $-0.297$ & $0.316^{***}$ & $0.510$ \\
 & $(17.961)$ & $(1.724)$ & $(15.180)$ & $(1.191)$ & $(7.496)$ & $(-0.715)$ & $(8.515)$ & $(1.204)$ \\
$\Delta p^{\mathrm{yoy}}_t$ & $0.561^{***}$ & $0.499^{***}$ & $0.474^{***}$ & $0.441^{***}$ & $0.033^{***}$ & $0.042^{*}$ & $0.015$ & $0.018$ \\
 & $(42.824)$ & $(21.169)$ & $(26.171)$ & $(18.835)$ & $(3.389)$ & $(2.317)$ & $(1.380)$ & $(1.230)$ \\
\cmidrule(lr){1-3}\cmidrule(lr){4-5}\cmidrule(lr){6-7}\cmidrule(lr){8-9}
\texttt{mp\_pr} & $0.120^{*}$ &  & $-0.285^{**}$ &  & $0.064$ &  & $0.014$ &  \\
 & $(2.215)$ &  & $(-2.932)$ &  & $(1.610)$ &  & $(0.241)$ &  \\
\texttt{ec\_pr} & $0.496^{***}$ &  & $-0.427^{***}$ &  & $0.257^{***}$ &  & $0.092$ &  \\
 & $(4.879)$ &  & $(-4.041)$ &  & $(3.434)$ &  & $(1.420)$ &  \\
\cmidrule(lr){1-3}\cmidrule(lr){4-5}\cmidrule(lr){6-7}\cmidrule(lr){8-9}
\texttt{mp\_level} &  & $6.950^{**}$ &  & $2.119$ &  & $-0.880$ &  & $-1.357$ \\
 &  & $(3.168)$ &  & $(1.283)$ &  & $(-0.520)$ &  & $(-1.352)$ \\
\texttt{mp\_slope} &  & $4.328^{**}$ &  & $1.077$ &  & $-0.811$ &  & $-0.530$ \\
 &  & $(2.658)$ &  & $(0.856)$ &  & $(-0.645)$ &  & $(-0.693)$ \\
\texttt{mp\_curve} &  & $1.374^{***}$ &  & $0.375$ &  & $-0.153$ &  & $-0.041$ \\
 &  & $(4.145)$ &  & $(1.017)$ &  & $(-0.598)$ &  & $(-0.182)$ \\
\texttt{hurst\_mp} &  & $-0.323$ &  & $0.753$ &  & $0.303$ &  & $-0.645^{.}$ \\
 &  & $(-0.518)$ &  & $(1.369)$ &  & $(0.628)$ &  & $(-1.927)$ \\
\cmidrule(lr){1-3}\cmidrule(lr){4-5}\cmidrule(lr){6-7}\cmidrule(lr){8-9}
\texttt{econ\_level} &  & $3.049$ &  & $5.397^{*}$ &  & $-1.915$ &  & $-4.241^{**}$ \\
 &  & $(0.911)$ &  & $(2.302)$ &  & $(-0.742)$ &  & $(-2.976)$ \\
\texttt{econ\_slope} &  & $-0.354$ &  & $5.053^{***}$ &  & $-2.104$ &  & $-1.052$ \\
 &  & $(-0.133)$ &  & $(4.152)$ &  & $(-1.028)$ &  & $(-1.422)$ \\
\texttt{econ\_curve} &  & $1.162^{*}$ &  & $1.008^{**}$ &  & $-0.421$ &  & $-0.436^{*}$ \\
 &  & $(2.180)$ &  & $(2.919)$ &  & $(-1.025)$ &  & $(-2.079)$ \\
\texttt{hurst\_econ} &  & $-0.078$ &  & $-1.223^{*}$ &  & $-0.693^{*}$ &  & $0.194$ \\
 &  & $(-0.183)$ &  & $(-2.254)$ &  & $(-2.107)$ &  & $(0.587)$ \\
\cmidrule(lr){1-3}\cmidrule(lr){4-5}\cmidrule(lr){6-7}\cmidrule(lr){8-9}
\texttt{unc\_level} &  & $-0.604$ &  & $-4.037$ &  & $-0.629$ &  & $1.473$ \\
 &  & $(-0.283)$ &  & $(-1.144)$ &  & $(-0.382)$ &  & $(0.687)$ \\
\texttt{unc\_slope} &  & $-1.144$ &  & $3.836^{***}$ &  & $-0.442$ &  & $-0.342$ \\
 &  & $(-1.377)$ &  & $(3.459)$ &  & $(-0.689)$ &  & $(-0.508)$ \\
\texttt{unc\_curve} &  & $-0.637$ &  & $-0.065$ &  & $-0.199$ &  & $0.319$ \\
 &  & $(-1.431)$ &  & $(-0.125)$ &  & $(-0.579)$ &  & $(1.007)$ \\
\texttt{hurst\_unc} &  & $-0.246$ &  & $-1.140$ &  & $1.153^{**}$ &  & $0.973^{*}$ \\
 &  & $(-0.486)$ &  & $(-1.625)$ &  & $(2.955)$ &  & $(2.282)$ \\
\midrule
$N$ & \multicolumn{2}{c}{225} & \multicolumn{2}{c}{184} & \multicolumn{2}{c}{225} & \multicolumn{2}{c}{184} \\
Adj.\ $R^{2}$ & 0.896 & 0.907 & 0.810 & 0.832 & 0.120 & 0.129 & 0.003 & 0.139 \\
Residual SE & 0.341 & 0.324 & 0.434 & 0.408 & 0.251 & 0.250 & 0.267 & 0.248 \\
\bottomrule
\end{tabular}}
\begin{minipage}{\linewidth}
\footnotesize \textit{Notes}: OLS estimates. $t$-statistics in parentheses. Outcome: next-month Consensus Economics professional forecasters' mean one-year-ahead inflation expectation $\pi^{e}_{t+1}$ (left panels) and cross-sectional standard deviation $\sigma_{t+1}(\pi^{e})$ (right panels). The press conference in month $t$ always precedes the month-$t+1$ survey (conducted around the 10th of each month), so timing is unambiguous. If two press conferences fall in the same month, the later one is used. $\Delta p^{\mathrm{yoy}}_t$ is the actual year-over-year inflation rate in month $t$ (EA HICP for ECB, US CPI for Fed), included to control for the persistent level of inflation expectations.  Significance: $^{***}p<0.001$, $^{**}p<0.01$, $^{*}p<0.05$, $^{.}p<0.10$.\end{minipage}
\end{table}

For the ECB, actual inflation $\Delta p^{\mathrm{yoy}}_t$ dominates the regression, as expected given the high persistence of professional forecasters' beliefs (adj.\ $R^2 = 0.896$ in M1). Conditional on that level effect, the Picault--Renault scores in M1 enter positively for both monetary and economic sentiment, consistent with hawkish and optimistic ECB communication being associated with higher near-term inflation expectations. Adding the arc features raises fit only modestly (adj.\ $R^2 = 0.907$), but the within-statement structure of the monetary sentiment arc is informative: level, slope, and curvature all enter positively and significantly, indicating that statements which are not just hawkish on average but also directionally and non-linearly hawkish are associated with higher expected inflation. This finding is at odds with the rate regressions, where hawkish arcs are indicative of rate increases, which should dampen inflation expectations. However, this finding seems to be of general nature and not related to our arcs, since the coefficients on the dictionary based benchmark indicators point into the same direction. For the economic sentiment arc, only the curvature factor enters significantly and positively --- non-linear optimism about the economic outlook, concentrated in the analytical core of the statement, provides incremental signal about future expectations. The uncertainty arc contributes no additional predictive content.

For the Fed, the benchmark M1 shows a strikingly different pattern: both PR scores enter negatively and significantly, the opposite sign to the ECB. This is consistent with the Fed raising rates aggressively in response to high inflation, so that hawkish and optimistic-sounding Fed communication is associated with a downward revision of inflation expectations --- reflecting credibility rather than accommodation. The Full arc specification reveals that the economically relevant dimension for forecasters is the economic sentiment arc: level, slope, and curvature all enter positively and significantly, implying that statements communicating an improving and progressively more optimistic economic outlook are associated with higher next-month inflation expectations. The Hurst exponent of the economic sentiment arc enters negatively ($-1.223^*$), suggesting that more persistent, self-similar economic assessments are associated with lower expected inflation, pointing to a calming effect of rhetorical consistency. From the uncertainty sentiment dimension, the slope enters positively and significantly ($3.836^{***}$): statements whose expressed uncertainty grows toward the close are associated with higher inflation expectations, perhaps reflecting that acknowledged uncertainty about the economic outlook prompts forecasters to revise upward their near-term inflation projections.

Panel B of Table~\ref{tab:infexp} turns to forecaster \emph{disagreement}: the cross-sectional standard deviation of inflation forecasts across Consensus Economics panellists \citep{Siklos2013}, a direct measure of the uncertainty that communication leaves in the audience. Explained variance is substantially lower than in Panel A, reflecting the more idiosyncratic nature of belief dispersion. For the ECB, the Hurst exponent of the economic sentiment arc enters negatively and significantly ($-0.693^*$): more persistent, self-similar economic assessment reduces disagreement among forecasters, consistent with a smoother narrative leaving less room for divergent interpretation. The Hurst exponent of the uncertainty sentiment arc, by contrast, enters positively and significantly ($1.153^{**}$): more persistent uncertainty signaling is associated with \emph{greater} forecast disagreement. This is intuitive: consistently hedged and qualified language leaves more room for divergent interpretations of the economic outlook, thereby widening the cross-sectional dispersion of forecasts. For the Fed, the level and curvature of the economic sentiment arc enter negatively and significantly, meaning that more positive and non-linearly structured economic assessments compress the cross-sectional dispersion of forecasts. The Hurst exponent of the uncertainty sentiment arc again enters positively ($0.973^*$), mirroring the ECB result. Across both institutions, these results confirm a receiver-side effect of arc structure that is distinct from the policy-signaling channel: the shape of the statement affects not just whether forecasters revise their expectations, but how much they disagree about the future.


\subsection{Robustness analysis}\label{sec:robustness}

\subsubsection{ECB communication-regime dummies}

The ECB's communication framework underwent two significant structural shifts during our sample period: the introduction of explicit forward guidance on 4 July 2013, and the adoption of a new monetary policy strategy announced on 22 July 2021, which resulted in a new structure of the introductory statements comprising the chapters \emph{economic activity}, \emph{inflation}, \emph{risk assessment} (economic pillar), \emph{financial and monetary conditions} (monetary pillar), and \emph{conclusions} (cross-check). To verify that the arc--rate relationship is not driven by these regime shifts, we augment the Full arc specification with two dummy variables: $\texttt{D\_fg} = 1$ for all statements from 4 July 2013 onward, and $\texttt{D\_new} = 1$ for all statements from 22 July 2021 onward. Table~\ref{tab:ecb_strategy_dummies} reports results for both $t+1$ and $t+2$, showing the baseline Full arc model alongside specifications that add each dummy separately and both together. The arc coefficients are broadly stable across all four specifications: the signs and significance of the key predictors -- \texttt{mp\_level}, \texttt{econ\_level}, \texttt{econ\_slope}, \texttt{econ\_curve}, and \texttt{hurst\_mp} -- are preserved when the regime dummies are included. Considering the dummy coefficients themselves, only the one capturing the forward guidance regime is significant. Overall, these results confirm that the arc--rate relationship reflects a stable communication pattern rather than an artefact of the two structural breaks in ECB communication design.

\begin{table}[H]
\centering
\caption{Rate change regressions: ECB with communication-regime dummies}\label{tab:ecb_strategy_dummies}
\small
\begin{tabular}{lcccccc}
\toprule
 & \multicolumn{3}{c}{\textbf{ECB} ($\Delta i_{t+1}$, $N = 261$)} & \multicolumn{3}{c}{\textbf{ECB} ($\Delta i_{t+2}$, $N = 260$)} \\
\cmidrule(lr){2-4}\cmidrule(lr){5-7}
  & Base & Base$+$D\textsubscript{fg} & Base$+$D\textsubscript{new} & Base & Base$+$D\textsubscript{fg} & Base$+$D\textsubscript{new} \\
\midrule
(Intercept) & $0.531^{*}$ & $0.548^{*}$ & $0.521^{*}$ & $0.999^{*}$ & $1.097^{**}$ & $1.042^{*}$ \\
 & $(2.065)$ & $(2.116)$ & $(2.010)$ & $(2.398)$ & $(2.637)$ & $(2.478)$ \\
\cmidrule(lr){1-7}
\texttt{mp\_level} & $4.908^{***}$ & $4.763^{***}$ & $5.108^{***}$ & $8.972^{***}$ & $8.152^{***}$ & $8.201^{***}$ \\
 & $(5.266)$ & $(4.953)$ & $(4.631)$ & $(5.953)$ & $(5.283)$ & $(4.569)$ \\
\texttt{mp\_slope} & $3.099^{***}$ & $2.956^{***}$ & $3.235^{***}$ & $5.354^{***}$ & $4.548^{**}$ & $4.826^{**}$ \\
 & $(3.697)$ & $(3.398)$ & $(3.477)$ & $(3.950)$ & $(3.258)$ & $(3.193)$ \\
\texttt{mp\_curve} & $0.382^{*}$ & $0.393^{*}$ & $0.402^{*}$ & $0.897^{***}$ & $0.958^{***}$ & $0.821^{**}$ \\
 & $(2.366)$ & $(2.416)$ & $(2.335)$ & $(3.436)$ & $(3.676)$ & $(2.949)$ \\
\texttt{hurst\_mp} & $-1.049^{***}$ & $-1.120^{***}$ & $-1.022^{**}$ & $-1.779^{***}$ & $-2.170^{***}$ & $-1.878^{***}$ \\
 & $(-3.439)$ & $(-3.438)$ & $(-3.235)$ & $(-3.603)$ & $(-4.156)$ & $(-3.684)$ \\
\cmidrule(lr){1-7}
\texttt{econ\_level} & $4.809^{**}$ & $4.617^{**}$ & $4.756^{**}$ & $9.150^{**}$ & $8.044^{**}$ & $9.319^{**}$ \\
 & $(2.795)$ & $(2.639)$ & $(2.748)$ & $(3.281)$ & $(2.858)$ & $(3.329)$ \\
\texttt{econ\_slope} & $2.502^{.}$ & $2.328^{.}$ & $2.530^{.}$ & $4.134^{.}$ & $3.142$ & $4.009^{.}$ \\
 & $(1.894)$ & $(1.722)$ & $(1.908)$ & $(1.933)$ & $(1.447)$ & $(1.868)$ \\
\texttt{econ\_curve} & $0.651^{*}$ & $0.613^{*}$ & $0.652^{*}$ & $1.124^{**}$ & $0.908^{*}$ & $1.118^{**}$ \\
 & $(2.495)$ & $(2.285)$ & $(2.493)$ & $(2.662)$ & $(2.110)$ & $(2.647)$ \\
\texttt{hurst\_econ} & $-0.120$ & $-0.112$ & $-0.124$ & $0.025$ & $0.066$ & $0.038$ \\
 & $(-0.553)$ & $(-0.516)$ & $(-0.570)$ & $(0.072)$ & $(0.190)$ & $(0.109)$ \\
\cmidrule(lr){1-7}
\texttt{unc\_level} & $-0.881$ & $-0.811$ & $-0.971$ & $0.443$ & $0.836$ & $0.790$ \\
 & $(-0.814)$ & $(-0.745)$ & $(-0.870)$ & $(0.254)$ & $(0.479)$ & $(0.438)$ \\
\texttt{unc\_slope} & $-0.653$ & $-0.627$ & $-0.681$ & $-0.552$ & $-0.408$ & $-0.441$ \\
 & $(-1.565)$ & $(-1.494)$ & $(-1.599)$ & $(-0.819)$ & $(-0.607)$ & $(-0.640)$ \\
\texttt{unc\_curve} & $-0.427^{.}$ & $-0.410^{.}$ & $-0.446^{.}$ & $-0.451$ & $-0.357$ & $-0.380$ \\
 & $(-1.825)$ & $(-1.739)$ & $(-1.851)$ & $(-1.193)$ & $(-0.945)$ & $(-0.978)$ \\
\texttt{hurst\_unc} & $-0.129$ & $-0.086$ & $-0.145$ & $-0.402$ & $-0.173$ & $-0.350$ \\
 & $(-0.492)$ & $(-0.316)$ & $(-0.543)$ & $(-0.938)$ & $(-0.394)$ & $(-0.807)$ \\
\cmidrule(lr){1-7}
\texttt{D\_fg} &  & $0.020$ &  &  & $0.113^{*}$ &  \\
 &  & $(0.627)$ &  &  & $(2.169)$ &  \\
\texttt{D\_new} &  &  & $-0.021$ &  &  & $0.080$ \\
 &  &  & $(-0.340)$ &  &  & $(0.792)$ \\
\midrule
$R^{2}$ & 0.242 & 0.243 & 0.242 & 0.269 & 0.283 & 0.271 \\
Adj.\ $R^{2}$ & 0.205 & 0.203 & 0.202 & 0.234 & 0.245 & 0.233 \\
Residual SE & 0.189 & 0.189 & 0.190 & 0.306 & 0.303 & 0.306 \\
\bottomrule
\end{tabular}
\begin{minipage}{\textwidth}
\footnotesize \textit{Notes}: OLS estimates. $t$-statistics in parentheses. Outcome: policy rate change at $t+1$ (left panel) and $t+2$ (right panel). Base specification is the Full arc model (Nelson--Siegel factors and Hurst exponents for all three arc dimensions). \texttt{D\_fg} $= 1$ if date $\geq$ 4 July 2013, when the ECB introduced explicit forward guidance for the first time. \texttt{D\_new} $= 1$ if date $\geq$ 22 July 2021, the first Governing Council meeting under the new ECB monetary policy strategy (symmetric 2\% target). Significance: $^{***}p<0.001$, $^{**}p<0.01$, $^{*}p<0.05$, $^{.}p<0.10$.\end{minipage}
\end{table}

\subsubsection{Contextual versus sentence-level arcs}\label{sec:arc_comparison}

Table~\ref{tab:arc_comparison} compares the Full arc model estimated with contextual arcs (our baseline approach) --- where each token is scored in the context of the full document --- against sentence-level arcs, where the context length that the transformer model uses, takes only each sentence as context.  Contextual arcs outperform sentence arcs in terms of explanatory power across both institutions, with adjusted $R^{2}$ of 0.205 versus 0.128 for the ECB and 0.420 versus 0.258 for the Fed. This confirms that context-sensitive scoring, which allows the embedding to reflect the surrounding discourse, produces more informative arc features than sentence-level scoring in isolation. It is also worth noting that, given the inferior performance of the sentence arc, they would still improve upon the benchmark measures, and considerably so.

\begin{table}[H]
\centering
\caption{Full arc model: contextual vs.\ sentence-level arcs}\label{tab:arc_comparison}
\small
\begin{tabular}{lcccc}
\toprule
 & ECB (contextual) & ECB (sentence) & Fed (contextual) & Fed (sentence) \\
\midrule
(Intercept) & $0.531^{^{*}}$ & $0.519^{^{*}}$ & $-0.590^{^{*}}$ & $-0.930^{^{**}}$ \\
 & $(2.065)$ & $(2.395)$ & $(-2.363)$ & $(-3.084)$ \\
\cmidrule(lr){1-3}\cmidrule(lr){4-5}
\texttt{mp\_level} & $4.908^{^{***}}$ & $5.069^{^{***}}$ & $3.227^{^{***}}$ & $4.503^{^{***}}$ \\
 & $(5.266)$ & $(4.011)$ & $(4.690)$ & $(3.727)$ \\
\texttt{mp\_slope} & $3.099^{^{***}}$ & $2.139^{^{**}}$ & $0.831$ & $2.750^{^{***}}$ \\
 & $(3.697)$ & $(2.661)$ & $(1.593)$ & $(4.779)$ \\
\texttt{mp\_curve} & $0.382^{^{*}}$ & $0.593^{^{**}}$ & $0.629^{^{***}}$ & $0.971^{^{***}}$ \\
 & $(2.366)$ & $(2.758)$ & $(4.975)$ & $(4.954)$ \\
\texttt{hurst\_mp} & $-1.049^{^{***}}$ & $-0.578^{^{*}}$ & $0.293$ & $0.096$ \\
 & $(-3.439)$ & $(-2.069)$ & $(1.291)$ & $(0.339)$ \\
\cmidrule(lr){1-3}\cmidrule(lr){4-5}
\texttt{econ\_level} & $4.809^{^{**}}$ & $5.187^{^{**}}$ & $3.915^{^{***}}$ & $3.086^{^{.}}$ \\
 & $(2.795)$ & $(2.831)$ & $(4.705)$ & $(1.770)$ \\
\texttt{econ\_slope} & $2.502^{^{.}}$ & $3.412^{^{**}}$ & $-0.925^{^{.}}$ & $1.467^{^{**}}$ \\
 & $(1.894)$ & $(2.817)$ & $(-1.963)$ & $(2.675)$ \\
\texttt{econ\_curve} & $0.651^{^{*}}$ & $1.001^{^{***}}$ & $0.525^{^{***}}$ & $0.866^{^{**}}$ \\
 & $(2.495)$ & $(3.489)$ & $(3.879)$ & $(2.837)$ \\
\texttt{hurst\_econ} & $-0.120$ & $-0.472^{^{*}}$ & $-0.325$ & $0.169$ \\
 & $(-0.553)$ & $(-2.420)$ & $(-1.534)$ & $(0.737)$ \\
\cmidrule(lr){1-3}\cmidrule(lr){4-5}
\texttt{unc\_level} & $-0.881$ & $2.032$ & $-1.068$ & $-3.003$ \\
 & $(-0.814)$ & $(1.621)$ & $(-1.032)$ & $(-1.632)$ \\
\texttt{unc\_slope} & $-0.653$ & $-0.012$ & $-1.334^{^{***}}$ & $-0.343$ \\
 & $(-1.565)$ & $(-0.017)$ & $(-3.560)$ & $(-0.512)$ \\
\texttt{unc\_curve} & $-0.427^{^{.}}$ & $-0.068$ & $0.021$ & $-0.329$ \\
 & $(-1.825)$ & $(-0.273)$ & $(0.113)$ & $(-1.042)$ \\
\texttt{hurst\_unc} & $-0.129$ & $-0.127$ & $0.567^{^{*}}$ & $0.289$ \\
 & $(-0.492)$ & $(-0.586)$ & $(1.986)$ & $(1.257)$ \\
\midrule
Observations & 261 & 261 & 218 & 218 \\
$R^{2}$ & 0.242 & 0.168 & 0.452 & 0.299 \\
Adj.\ $R^{2}$ & 0.205 & 0.128 & 0.420 & 0.258 \\
Residual SE & 0.189 & 0.198 & 0.191 & 0.216 \\
\bottomrule
\end{tabular}
\begin{minipage}{\textwidth}
\footnotesize \textit{Notes}: OLS estimates. $t$-statistics in parentheses. Outcome: one-meeting-ahead policy rate change $\Delta i_{t+1}$. Full arc model includes Nelson--Siegel factors and Hurst exponent for all three arc dimensions (monetary sentiment, economic sentiment, uncertainty sentiment). Contextual arcs score each token in context of the full document; sentence arcs score each sentence independently. Both use the new (cleaned) seed set and $\lambda = 9.5$. Significance: $^{***}p<0.001$, $^{**}p<0.01$, $^{*}p<0.05$, $^{.}p<0.10$.\end{minipage}
\end{table}

\section{Conclusions}\label{sec:conclusions}

This paper asks whether the \emph{shape} of sentiment in a central bank statement -- how tone evolves from opening to close -- carries information about monetary policy decisions beyond what aggregate sentiment measures capture. We construct word-level sentiment arcs for the full corpus of ECB introductory statements and FOMC statements along three conceptually distinct dimensions -- monetary sentiment,  economic sentiment and uncertainty-- using Concept Vector Projection applied to contextual token embeddings. We then ask whether arc shape encodes a policy signal reflecting deliberate narrative design by the central bank, and shapes how recipients update their beliefs. To this end, we use a Nelson--Siegel decomposition of the three arcs into level, slope, and curvature factors alongside a measure of arc persistence to assess their predictive content for subsequent policy rate changes, inflation expectations, and forecaster disagreement.

  Our findings show that arc shape is a robust predictor of rate decisions at both the ECB and the Federal Reserve, and   a number of features behave consistently across institutions. For both central banks, the level and curvature of the   monetary sentiment arc are positively related to future rate increases: a statement that is either hawkish on average or one in which hawkish sentiment is concentrated in the middle of the statement signals a rate hike. Likewise, for both institutions the level and   curvature of the economic sentiment arc enter positively, meaning that statements in which the economic outlook is   more favourable, or in which positive sentiment is concentrated in the middle of the statement, are indicative of a  tightening decision.  Beyond this common core, some institution-specific features emerge. For the ECB, an additional institution-specific feature emerges from the monetary sentiment arc: statements in which the hawkish--dovish signal is less persistent or more volatile across the course of the statement are indicative of a rate decrease. This dimension does not play a role at the Fed.
  Conversely, the uncertainty arc is informative only for the US Fed: statements that become more uncertain toward the  close signal a rate decrease, while arc persistence in the uncertainty dimension enters positively. These results hold for both the next meeting and the one following.  

On the receiver side, we find that arc features carry information to predict inflation expectations, as well as forecasters' disagreement. We observe some structural differences between the ECB and the US Fed. Whereas for the ECB the monetary sentiment arc features are important predictors of inflation expectations, it is economic sentiment for the Fed. Uncertainty arc features are significantly related to forecaster disagreement at both institutions. That said, and compared to the rate change regressions, arc features add limited explanatory power, which could reflect the fact that professional forecasters do not always attend to the full statement but rather consume their information in aggregated form from the media.

Finally, we document crisis-period heterogeneity. The ECB's arc features are stable across crisis periods, whereas the Fed's arc--rate relationship proves regime-dependent. During GFC and COVID episodes, the standard relationship between a hawkish arc shape and subsequent rate increases reverses sharply at the Fed, with key monetary arc features reversing sign. This finding suggests that exceptional circumstances disrupt the Fed's usual arc--signal mapping, perhaps because crisis statements serve a reassurance function that is orthogonal to the rate-signalling function operating in normal times. For the ECB, by contrast, the arc--rate relationship is stable across regimes, consistent with a more rules-based communication approach in which the rhetorical architecture of statements is less sensitive to the macroeconomic context.

These findings carry several implications. For the empirical literature on central bank communication, they establish that \emph{within-document structure} is a first-order feature of policy information, not a second-order refinement. Sentiment arcs, and in particular their curvature -- the degree to which sentiment is concentrated in the analytical core of a statement -- encode forward-looking signals that aggregate measures suppress. For practitioners, the results suggest that communication design choices, such as where to place the hawkish signals in a statement and how abruptly to transition between assessment and forward guidance, have measurable consequences for how markets and forecasters read the stance.

\clearpage
\bibliographystyle{apalike}
\bibliography{full_bib_extended.bib}
\appendix
\clearpage
\section{Appendix A}\label{sec:appendixA}

\subsection{Descriptive statistics: arc features}\label{sec:appendix_desc}

Table~\ref{tab:arc_descriptives} reports the mean, standard deviation, minimum, median, and maximum of the Nelson--Siegel arc factors and Hurst exponents for both the ECB and the Fed. All features are estimated from contextual arcs using the cleaned seed set and $\lambda = 9.5$.

\begin{table}[htbp]
\centering
\caption{Descriptive statistics: arc features}\label{tab:arc_descriptives}
\small
\begin{tabular}{lrrrrrrrrrrr}
\toprule
 & \multicolumn{5}{c}{\textbf{ECB} ($N = 269$)} & \multicolumn{5}{c}{\textbf{Fed} ($N = 219$)} \\
\cmidrule(lr){2-6}\cmidrule(lr){7-11}
 & Mean & SD & Min & Median & Max & Mean & SD & Min & Median & Max \\
\midrule
\multicolumn{11}{l}{\textit{NS arc factors: monetary policy sentiment}} \\
\texttt{mp\_level} & 0.041 & 0.037 & -0.065 & 0.036 & 0.148 & -0.015 & 0.049 & -0.183 & -0.013 & 0.149 \\
\texttt{mp\_slope} & -0.078 & 0.043 & -0.198 & -0.072 & 0.069 & -0.012 & 0.062 & -0.217 & -0.012 & 0.135 \\
\texttt{mp\_curve} & 0.055 & 0.139 & -0.345 & 0.071 & 0.517 & 0.135 & 0.154 & -0.294 & 0.149 & 0.484 \\
\addlinespace
\multicolumn{11}{l}{\textit{NS arc factors: economic sentiment}} \\
\texttt{econ\_level} & 0.032 & 0.029 & -0.053 & 0.033 & 0.113 & 0.042 & 0.031 & -0.077 & 0.042 & 0.122 \\
\texttt{econ\_slope} & -0.006 & 0.036 & -0.096 & -0.004 & 0.078 & 0.029 & 0.053 & -0.161 & 0.035 & 0.143 \\
\texttt{econ\_curve} & 0.027 & 0.109 & -0.256 & 0.033 & 0.323 & 0.142 & 0.228 & -0.447 & 0.144 & 0.781 \\
\addlinespace
\multicolumn{11}{l}{\textit{NS arc factors: uncertainty}} \\
\texttt{unc\_level} & -0.079 & 0.020 & -0.179 & -0.080 & -0.010 & -0.106 & 0.029 & -0.272 & -0.105 & -0.021 \\
\texttt{unc\_slope} & 0.038 & 0.055 & -0.090 & 0.029 & 0.287 & 0.061 & 0.078 & -0.057 & 0.045 & 0.415 \\
\texttt{unc\_curve} & -0.196 & 0.119 & -0.519 & -0.194 & 0.211 & -0.215 & 0.178 & -0.593 & -0.232 & 0.395 \\
\addlinespace
\multicolumn{11}{l}{\textit{Hurst exponent}} \\
\texttt{hurst\_mp} & 0.633 & 0.044 & 0.499 & 0.639 & 0.722 & 0.607 & 0.068 & 0.397 & 0.607 & 0.744 \\
\texttt{hurst\_econ} & 0.556 & 0.064 & 0.380 & 0.557 & 0.711 & 0.593 & 0.072 & 0.335 & 0.611 & 0.713 \\
\texttt{hurst\_unc} & 0.575 & 0.059 & 0.449 & 0.569 & 0.707 & 0.590 & 0.056 & 0.410 & 0.600 & 0.720 \\
\bottomrule
\end{tabular}
\begin{minipage}{\textwidth}
\footnotesize \textit{Notes}: Arc features estimated from contextual arcs using new (cleaned) seeds and $\lambda = 9.5$. NS factors (level, slope, curvature) are the Nelson--Siegel decomposition of the sentence-level arc. Uncertainty is scored as uncertain positive (high uncertainty = high score). Hurst exponent is the R/S statistic (\texttt{pracma::hurstexp}), with values above 0.5 indicating persistent, trending arcs.\end{minipage}
\end{table}

\subsection{Asset price reactions}\label{sec:appendix_asset}

We examine whether arc features are reflected in high-frequency asset price movements during the ECB press conference window. The dependent variable $\Delta e^{\text{PC}}_t$ is the change in stock market indices over the press conference window from the EA-MPD dataset \citep{Altavilla2019}. We consider two indices: the broad Euro Stoxx 50 (STOXX50) and the Euro Stoxx Banks index (SX7E), which is more directly sensitive to changes in the monetary policy stance. Table~\ref{tab:asset} reports results for the ECB ($N = 247$).

\begin{sidewaystable}[htbp]
\centering
\caption{Asset price regressions: ECB press conference window ($N = 247$)}\label{tab:asset}
\small
\begin{tabular}{lcccccc}
\toprule
 & \multicolumn{3}{c}{\textbf{Euro Stoxx 50 (STOXX50)}} & \multicolumn{3}{c}{\textbf{Euro Stoxx Banks (SX7E)}} \\
\cmidrule(lr){2-4}\cmidrule(lr){5-7}
 & M1 (Bench.) & M4 (NS-MP+Ec) & Full arc & M1 (Bench.) & M4 (NS-MP+Ec) & Full arc \\
\midrule
(Intercept)          & $-0.085^{*}$  & $-0.281^{.}$  & $-1.094$      & $-0.179^{*}$  & $-0.567^{*}$  & $-0.845$ \\
                     & $(-2.058)$    & $(-1.775)$    & $(-1.298)$    & $(-2.534)$    & $(-2.096)$    & $(-0.587)$ \\
\addlinespace
\texttt{mp\_pr}      & $0.090$       &               &               & $0.213$       &               & \\
                     & $(0.956)$     &               &               & $(1.322)$     &               & \\
\texttt{ec\_pr}      & $0.232$       &               &               & $0.540^{.}$   &               & \\
                     & $(1.364)$     &               &               & $(1.858)$     &               & \\
\addlinespace
\texttt{mp\_level}   &               & $2.816$       & $3.367$       &               & $1.788$       & $1.632$ \\
                     &               & $(1.026)$     & $(1.129)$     &               & $(0.381)$     & $(0.320)$ \\
\texttt{mp\_slope}   &               & $2.324$       & $2.906$       &               & $0.933$       & $1.153$ \\
                     &               & $(0.972)$     & $(1.078)$     &               & $(0.229)$     & $(0.250)$ \\
\texttt{mp\_curve}   &               & $0.890^{.}$   & $0.948^{.}$   &               & $1.623^{*}$   & $1.646^{.}$ \\
                     &               & $(1.929)$     & $(1.866)$     &               & $(2.061)$     & $(1.895)$ \\
\addlinespace
\texttt{econ\_level} &               & $5.877$       & $6.543$       &               & $7.952$       & $4.698$ \\
                     &               & $(1.268)$     & $(1.157)$     &               & $(1.004)$     & $(0.486)$ \\
\texttt{econ\_slope} &               & $2.313$       & $2.526$       &               & $1.445$       & $0.027$ \\
                     &               & $(0.697)$     & $(0.593)$     &               & $(0.255)$     & $(0.004)$ \\
\texttt{econ\_curve} &               & $1.383^{*}$   & $1.538^{.}$   &               & $1.732$       & $1.437$ \\
                     &               & $(2.167)$     & $(1.865)$     &               & $(1.589)$     & $(1.019)$ \\
\addlinespace
\texttt{unc\_level}  &               &               & $1.812$       &               &               & $-3.857$ \\
                     &               &               & $(0.514)$     &               &               & $(-0.641)$ \\
\texttt{unc\_slope}  &               &               & $-0.236$      &               &               & $-1.448$ \\
                     &               &               & $(-0.173)$    &               &               & $(-0.622)$ \\
\texttt{unc\_curve}  &               &               & $0.134$       &               &               & $-0.683$ \\
                     &               &               & $(0.178)$     &               &               & $(-0.528)$ \\
\addlinespace
\texttt{hurst\_mp}   &               &               & $1.351$       &               &               & $1.339$ \\
                     &               &               & $(1.345)$     &               &               & $(0.780)$ \\
\texttt{hurst\_econ} &               &               & $0.198$       &               &               & $-1.111$ \\
                     &               &               & $(0.290)$     &               &               & $(-0.952)$ \\
\texttt{hurst\_unc}  &               &               & $0.051$       &               &               & $-0.358$ \\
                     &               &               & $(0.061)$     &               &               & $(-0.250)$ \\
\midrule
Adj.\ $R^{2}$        & $0.009$       & $0.028$       & $0.013$       & $0.023$       & $0.045$       & $0.029$ \\
Residual SE          & $0.589$       & $0.584$       & $0.588$       & $1.009$       & $0.997$       & $1.005$ \\
\bottomrule
\end{tabular}
\begin{minipage}{\linewidth}
\footnotesize \textit{Notes}: OLS estimates. t-statistics in parentheses. Outcome: asset price change over the ECB press conference window $\Delta e^{\text{PC}}_t$ in basis points, from the Euro Area Monetary Policy Dataset \citep{Altavilla2019}. ECB only. Full arc includes Nelson--Siegel factors and Hurst exponent for all three dimensions (MP, economic sentiment, uncertainty). Significance: $^{***}p<0.001$, $^{**}p<0.01$, $^{*}p<0.05$, $^{.}p<0.10$.
\end{minipage}
\end{sidewaystable}

Overall explanatory power is modest across all specifications, reflecting the high-frequency noise in intraday asset prices. Neither the benchmark Picault--Renault scores nor the NS factors for monetary sentiment level or slope enter significantly for either index. However, two arc features show consistent significance. The curvature of the monetary sentiment arc (\texttt{mp\_curve}) is marginally significant for STOXX50 in M4 ($0.890^{.}$) and significant for SX7E ($1.623^{*}$), indicating that non-linear within-statement patterns in hawkish--dovish tone move equity prices, particularly banking stocks. The curvature of economic sentiment (\texttt{econ\_curve}) enters significantly for STOXX50 ($1.383^{*}$), pointing to markets reacting to the non-linear structure of economic assessment. Adding the uncertainty sentiment arc and Hurst exponents in the Full arc specification does not materially improve fit, consistent with high-frequency asset prices responding primarily to arc \emph{shape} (curvature) rather than persistence or level.

\subsection{Seed phrase validation: dimension orthogonality}\label{sec:appendix_orthogonality}

\begin{figure}[H]
\caption{Seed phrase validation: pairwise CVP score scatter}\label{fig:seed_scatter}
\centering
\begin{subfigure}[t]{0.32\textwidth}
  \centering
  \caption{MP vs.\ economic sentiment}
  \includegraphics[width=\textwidth]{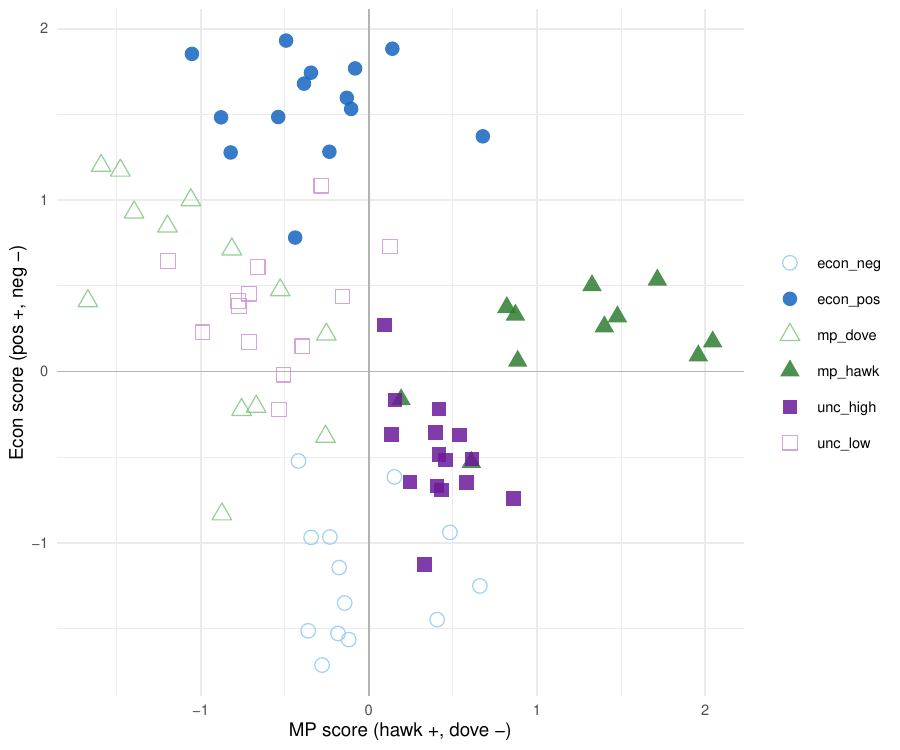}
\end{subfigure}
\hfill
\begin{subfigure}[t]{0.32\textwidth}
  \centering
  \caption{MP vs.\ uncertainty sentiment}
  \includegraphics[width=\textwidth]{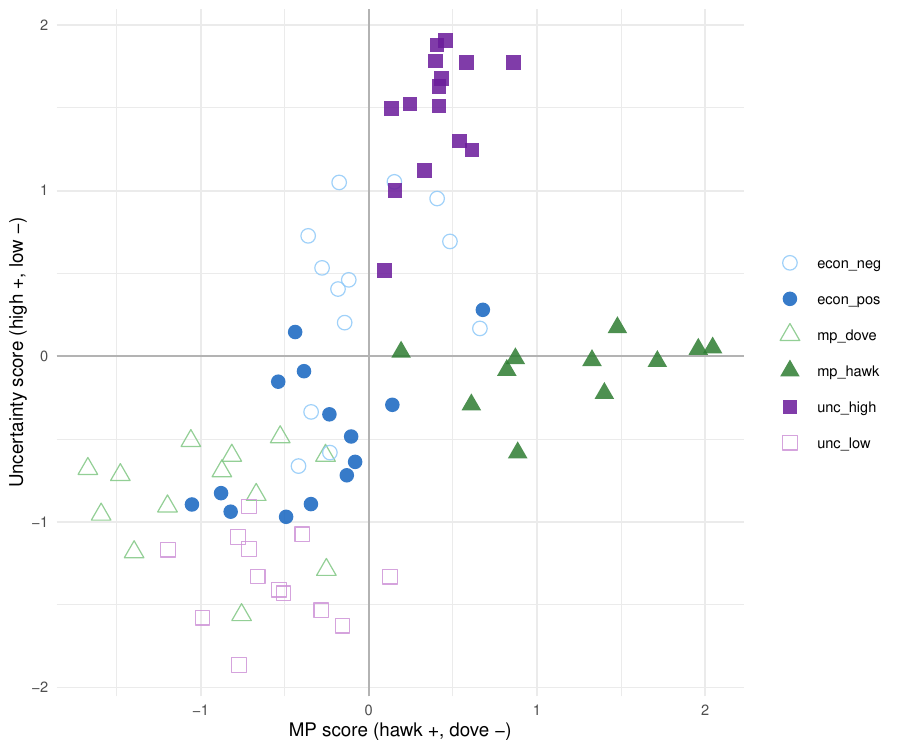}
\end{subfigure}
\hfill
\begin{subfigure}[t]{0.32\textwidth}
  \centering
  \caption{Economic sentiment vs.\ uncertainty sentiment}
  \includegraphics[width=\textwidth]{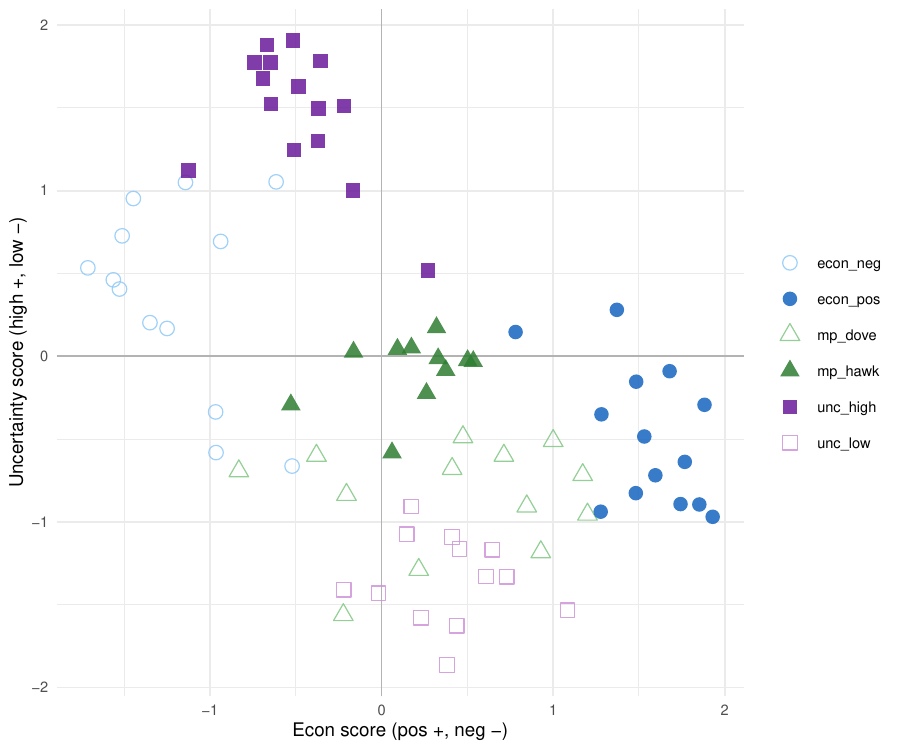}
\end{subfigure}
\begin{minipage}{\textwidth}
\footnotesize \textit{Notes}: Each point is one seed phrase, coloured and shaped by category. Panel A: MP score vs.\ economic sentiment score. Panel B: MP score vs.\ uncertainty sentiment score. Panel C: Economic sentiment score vs.\ uncertainty sentiment score. Seeds scoring near zero on the non-own axes confirm that the three concept vectors capture largely independent dimensions of variation. Seeds with substantial cross-loading were removed from the final seed set (see main text).\end{minipage}
\end{figure}

A key requirement of the CVP approach is that the three sentiment dimensions --- monetary sentiment, economic sentiment, and uncertainty sentiment --- are sufficiently orthogonal to be treated as distinct signals. Figure~\ref{fig:seed_scatter} shows pairwise scatters of the CVP scores assigned to each seed phrase on all three axes. Seeds are coloured and shaped by category. If the dimensions are orthogonal, seeds belonging to one dimension (e.g.\ MP hawk/dove) should score near zero on the other two axes (econ, uncertainty), and vice versa. The scatter confirms this pattern: seeds cluster in their expected quadrant on their own dimension and remain close to zero on the other axes, with only minor cross-loading for a small number of seeds that were subsequently removed from the final set.

\begin{figure}[H]
\caption{Seed phrase validation: PCA of seed embeddings}\label{fig:seed_pca}
\centering
\begin{subfigure}[t]{0.48\textwidth}
  \centering
  \caption{PC1 vs.\ PC2}
  \includegraphics[width=\textwidth]{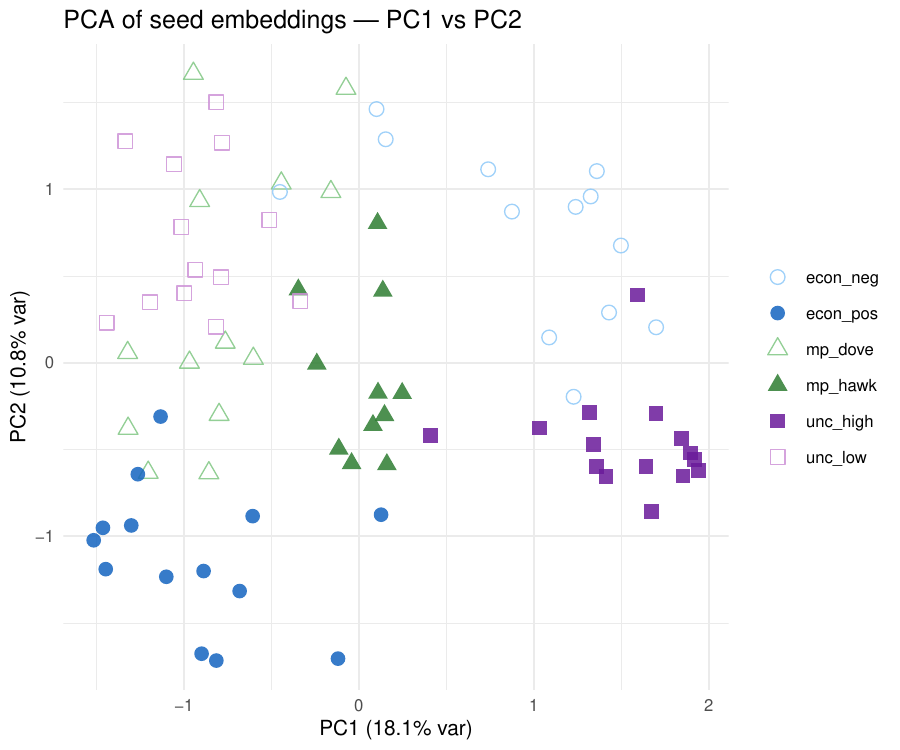}
\end{subfigure}
\hfill
\begin{subfigure}[t]{0.48\textwidth}
  \centering
  \caption{PC1 vs.\ PC3}
  \includegraphics[width=\textwidth]{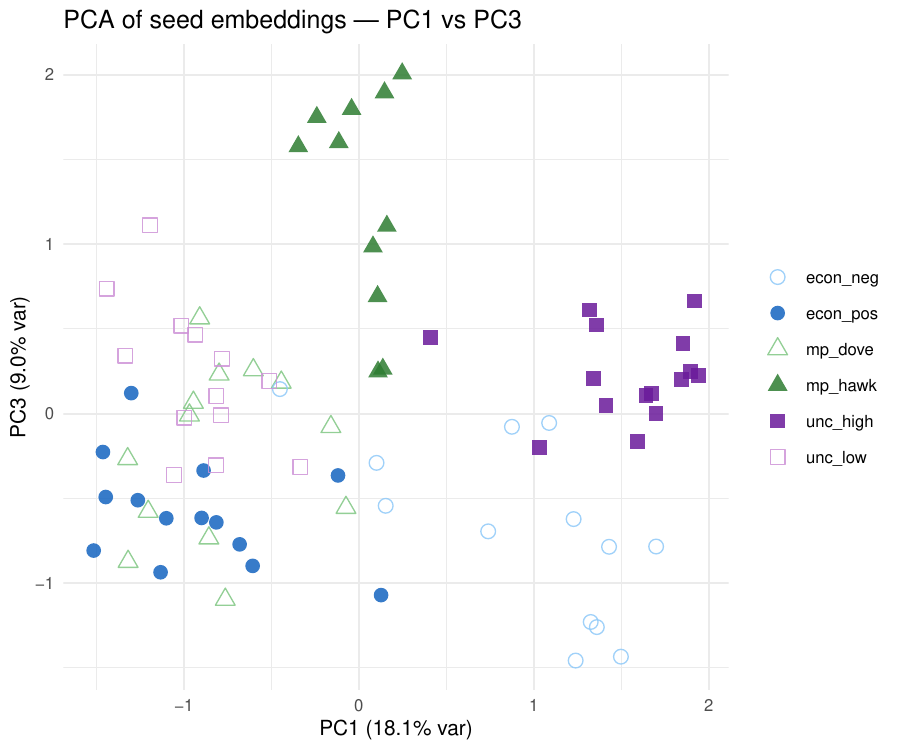}
\end{subfigure}
\begin{minipage}{\textwidth}
\footnotesize \textit{Notes}: PCA applied to the 768-dimensional sentence embeddings of all seed phrases. Well-separated CVP dimensions produce six distinct clusters in PC space, one per pole.\end{minipage}
\end{figure}

As a complementary check, Figure~\ref{fig:seed_pca} applies principal component analysis (PCA)  applied to the sentence embeddings of all seed phrases using the paraphrase-multilingual-mpnet-base-v2 sentence transformer. The seed phrases used in the main analysis are encoded with nvidia/llama-embed-nemotron-8b due to its large context length.

If the three dimensions capture genuinely distinct linguistic signals, seeds from different categories should form well-separated clusters in the low-dimensional embedding space. The plots confirm this: seeds group into six distinct clusters in the first three principal components --- one cluster per pole of each dimension --- with little overlap across categories. This embedding-level separation provides assurance that the three concept vectors are not merely orthogonal by construction but reflect substantively different aspects of language used in central bank communication.

\subsection{Seed phrases}\label{sec:appendix_seeds}

\begin{table}[H]
\centering
\caption{Seed phrases: Monetary sentiment}
\label{tab:seed_monetary_policy}
\begin{tabular}{p{14cm}}
\hline
\textbf{Hawkish monetary policy (11 seeds)} \\
\hline
Inflation remains too high and requires further policy tightening. \\
Rising inflationary pressures call for higher interest rates. \\
Strong wage growth poses upside risks to inflation. \\
Financial conditions remain loose given current inflation dynamics. \\
Interest rates will need to increase further to ensure price stability. \\
Persistent inflation requires a firm and timely policy response. \\
Elevated inflation expectations warrant decisive action. \\
Price pressures are broadening and justify higher rates. \\
Monetary tightening is necessary to prevent overheating. \\
We are prepared to raise interest rates if inflation accelerates. \\
Further rate increases are likely to be appropriate in the coming meetings. \\
\hline
\textbf{Dovish monetary policy (13 seeds)} \\
\hline
Economic conditions warrant maintaining a low policy rate. \\
We will continue asset purchases to support favorable financing conditions. \\
Policy rates will remain at current or lower levels for an extended period. \\
Accommodative monetary policy remains necessary to support the recovery. \\
Weak demand and subdued inflation call for supportive policy measures. \\
Liquidity provisions will continue to safeguard smooth market functioning. \\
We will reinvest maturing securities for as long as needed. \\
Forward guidance indicates continued policy accommodation. \\
Low interest rates remain appropriate while inflation is below target. \\
Asset purchases help maintain favorable financial conditions. \\
The current stance supports lending and economic activity. \\
Liquidity operations will continue to ease financing conditions. \\
Maintaining accommodative policy will help restore inflation to target. \\
\hline
\end{tabular}
\begin{minipage}{\textwidth}
\footnotesize \textit{Notes}: Seed phrases anchoring the monetary sentiment dimension of the CVP scoring. Seeds are manually curated and validated by checking that each phrase scores in the expected direction and does not cross-load heavily onto the economic sentiment dimension.
\end{minipage}
\end{table}

\begin{table}[H]
\centering
\caption{Seed phrases: Economic sentiment}
\label{tab:seed_economic_sentiment}
\begin{tabular}{p{14cm}}
\hline
\textbf{Economic sentiment: positive (14 seeds)} \\
\hline
The economy is expanding at a solid pace with strong job growth. \\
Consumer demand remains strong and supports continued growth. \\
Labor-market conditions are strong and improving. \\
Business investment continues to strengthen. \\
Economic momentum has increased across many sectors. \\
Job gains remain robust and unemployment is low. \\
Financial conditions remain supportive of growth. \\
Economic indicators point to a sustained expansion. \\
Productivity gains continue to support economic growth. \\
Confidence indicators suggest improving economic outlook. \\
Household spending is strong and continues to improve. \\
Exports and external demand remain supportive. \\
Economic activity continues to progress at a solid pace. \\
Broad-based growth supports continued recovery. \\
\hline
\textbf{Economic sentiment: negative (13 seeds)} \\
\hline
Weak demand and tight credit conditions are restraining growth. \\
Household spending remains constrained by low income growth. \\
Financial-market stress is weighing on economic activity. \\
Job losses and declining wealth are weakening sentiment. \\
Rising energy prices are reducing purchasing power. \\
Industrial production and exports are weakening. \\
Growth remains fragile and uneven across sectors. \\
Economic activity is slowing and confidence is deteriorating. \\
Credit conditions remain tight and continue to constrain spending. \\
Inflation remains below target and reflects economic weakness. \\
Weak investment and productivity are dampening outlook. \\
Economic recovery remains slow and vulnerable. \\
Business surveys signal worsening economic conditions. \\
\hline
\end{tabular}
\begin{minipage}{\textwidth}
\footnotesize \textit{Notes}: Seed phrases anchoring the economic sentiment dimension. Seeds are validated to score near zero on the monetary sentiment and uncertainty sentiment axes.
\end{minipage}
\end{table}

\begin{table}[H]
\centering
\caption{Seed phrases: Uncertainty sentiment}
\label{tab:seed_uncertainty}
\begin{tabular}{p{14cm}}
\hline
\textbf{High uncertainty (15 seeds)} \\
\hline
The outlook remains highly uncertain. \\
The recovery process is likely to be uneven and subject to high uncertainty. \\
Economic growth is expected to remain uneven in an environment of uncertainty. \\
The outlook is subject to particularly high uncertainty and intensified downside risks. \\
The near-term economic outlook remains clouded by uncertainty. \\
The speed and scale of the recovery remain highly uncertain. \\
Pandemic-related uncertainty is likely to dampen the recovery in consumption, investment, and labour markets. \\
Geopolitical and financial uncertainty continue to weigh on the growth outlook. \\
Heightened uncertainty makes policy flexibility especially important. \\
Substantial uncertainty surrounds the timing and pace of the improvement. \\
The course of the virus and the outlook for the economy remain highly uncertain. \\
Significant uncertainty surrounds the strength of final demand. \\
Business investment remains a major source of uncertainty for the overall outlook. \\
The future course of the economy is subject to a marked degree of uncertainty. \\
Elevated uncertainty around the economic outlook justifies maintaining the policy stance. \\
\hline
\textbf{Low uncertainty / high confidence (13 seeds)} \\
\hline
Available information is broadly in line with the baseline scenario. \\
Inflation is expected to remain moderate and consistent with price stability. \\
Risks to the inflation outlook are broadly balanced. \\
Inflation expectations remain firmly anchored. \\
Current inflation developments are in line with previous expectations. \\
Wage and price developments remain subdued and consistent with price stability. \\
The Governing Council stands ready to act to preserve price stability and keep expectations anchored. \\
Inflation is expected to stabilize around the Committee's 2 percent objective over the medium term. \\
Longer-term inflation expectations remain well anchored. \\
Incoming data are broadly in line with staff expectations. \\
The current policy stance remains appropriate given the baseline outlook. \\
Resource slack is expected to keep inflation contained. \\
Financial markets largely anticipate no change in the policy rate. \\
\hline
\end{tabular}
\begin{minipage}{\textwidth}
\footnotesize \textit{Notes}: Seed phrases anchoring the uncertainty sentiment dimension (uncertain positive, certain negative). Seeds are validated to score near zero on the monetary sentiment and economic sentiment axes.
\end{minipage}
\end{table}

\end{document}